\pdfoutput=1
\documentclass[11pt]{article}

\usepackage[preprint]{acl}

\usepackage{times}
\usepackage{latexsym}

\usepackage[T1]{fontenc}

\usepackage[utf8]{inputenc}

\usepackage{microtype}

\usepackage{makecell}
\usepackage{tcolorbox}

\usepackage{inconsolata}

\usepackage{graphicx}

\usepackage{lipsum} 
\usepackage{booktabs}
\usepackage{multirow}
\usepackage{graphicx}
\usepackage[normalem]{ulem}
\useunder{\uline}{\ul}{}
\usepackage{amsmath}
\usepackage{enumitem}
\usepackage{subcaption}
\usepackage{array}
\usepackage{tcolorbox}
\usepackage{breqn}

\newtheorem{problem}{Problem}

\usepackage{xspace}
\newcommand{\ours}{{AD-LLM}\xspace}

\newcommand{\add}[1]{{#1}}

\title{\ours: Benchmarking Large Language Models for Anomaly Detection}

\author{
    \textbf{Tiankai Yang\textsuperscript{1}$^{,*}$},
    \textbf{Yi Nian\textsuperscript{1}$^{,*}$},
    \textbf{Shawn Li\textsuperscript{1}},
    \textbf{Ruiyao Xu\textsuperscript{2}},
    \textbf{Yuangang Li\textsuperscript{1}},
    \textbf{Jiaqi Li\textsuperscript{1}},
\\
    \textbf{Zhuo Xiao\textsuperscript{1}},
    \textbf{Xiyang Hu\textsuperscript{3}},
    \textbf{Ryan Rossi\textsuperscript{4}},
    \textbf{Kaize Ding\textsuperscript{2}},
    \textbf{Xia Hu\textsuperscript{5}},
    \textbf{Yue Zhao\textsuperscript{1}}
\\
    \textsuperscript{1}University of Southern California,
    \textsuperscript{2}Northwestern University,
\\
    \textsuperscript{3}Arizona State University,
    \textsuperscript{4}Adobe Research,
    \textsuperscript{5}Rice University
\\
    \texttt{
        \{tiankaiy, li.li02, yuangang, jli77629, zhuoxiao, yzhao010\}@usc.edu,
    }
\\
    \texttt{
        ynian.4@gmail.com, ruiyaoxu2028@u.northwestern.edu, xiyanghu@asu.edu
    }
\\
    \texttt{
        ryrossi@adobe.com, kaize.ding@northwestern.edu, xia.hu@rice.edu
    }
}

\newcommand{\gitlink}{https://github.com/USC-FORTIS/AD-LLM}
\newcommand\nnfootnote[1]{%
  \begin{NoHyper}
  \renewcommand\thefootnote{}\footnote{#1}%
  \addtocounter{footnote}{-1}%
  \end{NoHyper}
}
\begin{document}
\setlength{\abovedisplayskip}{5pt}
\setlength{\belowdisplayskip}{5pt}
\setlength{\abovedisplayshortskip}{5pt}
\setlength{\belowdisplayshortskip}{5pt}
\maketitle

\nnfootnote{$^*$Equal contribution.}

\begin{abstract}
Anomaly detection (AD) is an important machine learning task with many real-world uses, including fraud detection, medical diagnosis, and industrial monitoring. 
Within natural language processing (NLP), AD helps detect issues like spam, misinformation, and unusual user activity. 
Although large language models (LLMs) have had a strong impact on tasks such as text generation and summarization, their potential in AD has not been studied enough. 
This paper introduces \ours, the first benchmark that evaluates how LLMs can help with NLP anomaly detection. We examine three key tasks: (\textit{i}) zero-shot detection, using LLMs' pre-trained knowledge to perform AD without task-specific training; 
(\textit{ii}) data augmentation, generating synthetic data and category descriptions to improve AD models; 
and (\textit{iii}) model selection, using LLMs to suggest unsupervised AD models. 
Through experiments with different datasets, we find that LLMs can work well in zero-shot AD, that carefully designed augmentation methods are useful, and that explaining model selection for specific datasets remains challenging. 
Based on these results, we outline six future research directions on LLMs for AD.

\end{abstract}
\section{Introduction}
Anomaly detection (AD) is an important topic in machine learning (ML) that identifies samples differing from the general distribution \cite{zhao2019pyod,liu2024pygod}. 
This ability is critical for many practical applications, such as fraud detection \cite{fraud}, medical diagnosis \cite{medical}, software engineering \cite{liicse}, and industrial system monitoring \cite{outspot}. 
Within natural language processing (NLP), AD is also important for finding unusual text instances, which is needed for detecting spam \cite{spam}, misinformation \cite{misinfo}, or unusual user behavior \cite{social}.

In the current era of large language models (LLMs), we ask how AD can make use of their capabilities and what the current level of integration looks like. 
While LLMs have brought large improvements to areas such as text generation, summarization, and translation, their possible benefits for AD, especially in NLP, have received some attention \cite{li2024multi,kaize_survey} but have not been studied in detail. 

\begin{figure*}[!t]
    \centering
    \includegraphics[width=1\linewidth]{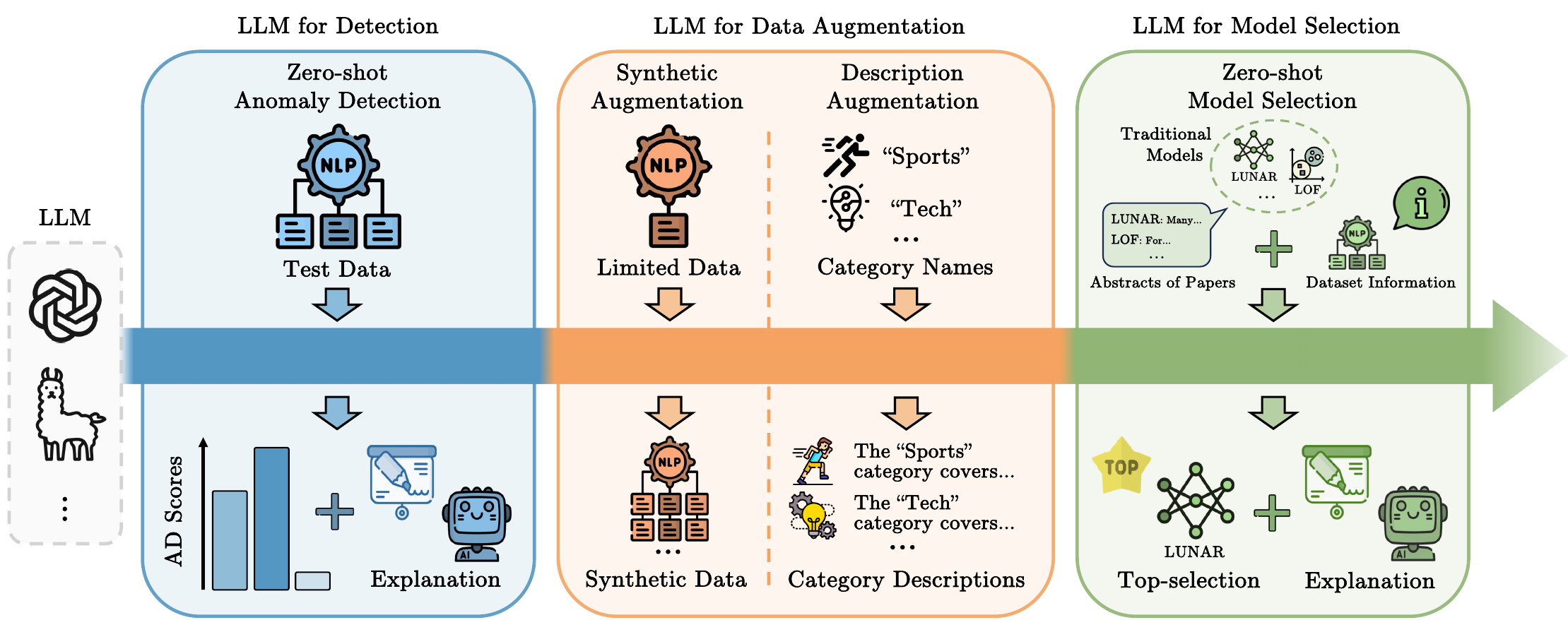}
    \caption{
    \ours examines 
    how LLMs contribute to three key AD tasks: 
    (\textbf{Task 1}, \S \ref{sec:ad}) Zero-shot detection (left), where LLMs directly identify anomalies and provide explanations without task-specific training data; 
    (\textbf{Task 2}, \S \ref{sec:aug}) Data augmentation (center), where LLMs generate synthetic samples and produce category descriptions to alleviate data scarcity and improve semantic reasoning; 
    and (\textbf{Task 3}, \S \ref{sec:ums}) Model selection (right), where LLMs analyze dataset attributes and model descriptions to recommend suitable AD models along with justifications. 
    }
    \label{fig:overview}
    \vspace{-10pt}
\end{figure*}


This work presents \textit{the first comprehensive benchmark}, called \ours, to study the roles and potential of LLMs in NLP anomaly detection. Our analysis focuses on three key tasks that are central in AD research and in practice (Figure  \ref{fig:overview}):

\begin{itemize}[nosep, itemsep=3pt, leftmargin=*]
     \item (\textit{i}) \textbf{LLM for Anomaly Detection} (\S \ref{sec:ad}): Many AD tasks lack enough labeled data, making it hard to train models from scratch \cite{han2022adbench}. 
    LLMs, with their pre-trained knowledge, can perform zero-shot AD \cite{kaize_survey}.


    \item (\textit{ii}) \textbf{LLM for Data Augmentation} (\S \ref{sec:aug}): AD tasks often suffer from unbalanced or limited data \cite{yoodata, limm}. 
    For example, only a few insurance fraud samples may be available \cite{bauder2018effects}.
    Generative LLMs may produce synthetic data to strengthen AD cost-effectively.
    \item (\textit{iii}) \textbf{LLM for Model Selection} (\S \ref{sec:ums}): Picking a good AD model usually needs many trials and domain insights \cite{jiang2024adgym}, and current choices in practice are often random \cite{zhao2021automatic}. 
    LLMs, with the prior knowledge and ability to reason, may be able to suggest suitable AD models and save human effort.
\end{itemize}

Collectively, these three tasks tackle fundamental AD challenges from multiple angles: rapidly detecting anomalies with minimal supervision, enriching limited datasets for more robust learning, 
and guiding model selection without extensive domain expertise. 
As a result, \ours not only improves individual AD components but also demonstrates \textit{how LLMs can streamline the entire process}—from raw data to reliable, actionable insights.

\noindent
\textbf{Key Takeaways.}  
Our results reveal several noteworthy insights: 
(\textit{i}) LLMs can achieve superior zero-shot AD performance, often outperforming conventional methods without relying on task-specific data. 
(\textit{ii}) Enriching LLM inputs with additional context, such as anomaly category names or descriptive prompts, further boosts detection quality.
(\textit{iii}) Employing LLM-driven data augmentation enhances AD performance, though the effectiveness varies with model features and dataset properties.
(\textit{iv}) LLM-based model selection can approach top-performing baselines, but improving interpretability and providing dataset-specific rationales remains an open area. 
These suggest future work that systematically integrates external knowledge, refines prompt engineering, and develops strategies to ensure more transparent, context-aware LLM recommendations in AD tasks.

\noindent
\textbf{Contributions.} This paper makes the following key contributions:
\begin{itemize}[nosep, itemsep=3pt, leftmargin=*]
    \item \textbf{The First Comprehensive LLM-based AD benchmark.}
    We introduce \ours, a unified evaluation framework that examines how LLMs address three core AD tasks—detection, data augmentation, and model selection.
    \item \textbf{Systematic and In-depth Experimental Analysis.}
    Through extensive experiments across multiple datasets, we show that LLMs can achieve strong zero-shot AD performance, boost AD methods by generating synthetic data or descriptive prompts, and recommend effective AD models w/o relying on historical performance data.
    \item \textbf{Reproducibility and Accessibility.}
    We release \ours under the MIT License at \url{\gitlink}, providing a platform for the community to explore advanced applications of LLMs in AD.
\end{itemize}

\section{Preliminaries on \ours}
\subsection{Related Work}
\label{sec:pre_related_work}


Recent studies have explored the role of LLMs in AD, highlighting both opportunities and challenges. 
\citet{kaize_survey} proposes a taxonomy categorizing LLMs as either detection or generative tools, but their work lacks experimental benchmarks. 
Similarly, \citet{jiang2024mmad} presents \textit{MMAD}, a benchmark designed for industrial AD, focusing on image datasets yet limiting its applicability to other modalities. 
\citet{liu-etal-2024-good} evaluates LLMs like Llama for out-of-distribution (OOD) detection, demonstrating the effectiveness of cosine distance detectors with isotropic embeddings achieved from LLMs. However, their study does not explore advanced LLM capabilities like data augmentation and zero-shot detection.

Our work, \textit{\ours}, bridges these gaps by introducing a comprehensive benchmark for evaluating LLMs in anomaly detection across diverse tasks. This makes \textit{\ours} a significant step toward advancing LLM-driven anomaly detection.


 






\subsection{Datasets and Traditional Baselines}
\vspace{-0.3ex}
\label{sec:pre_data_baseline}
Our experiments encompass five NLP AD datasets sourced from \citet{nlp_adbench}, derived from classification datasets. Each dataset contains text samples from multiple categories, with one designated as the anomaly category. The training data includes only normal samples. See the detailed information on datasets in Appx.~\ref{appx:pre_dataset}

We compare LLM-based AD with 18 traditional training-based unsupervised methods evaluated in \citet{nlp_adbench} and leverage LLMs to enhance them. These baselines can be categorized into two groups: (1) end-to-end algorithms that directly process raw text data to produce AD results and (2) two-step methods that first create text embeddings using language models and then apply traditional AD techniques to those embeddings. See a complete list of methods in Appx.~\ref{appx:pre_baseline}. 


\subsection{Common Experimental Settings}
\label{sec:pre_exp}
\vspace{-0.3ex}
\noindent
\textbf{Evaluation Metrics.}
We evaluate the AD performance using two commonly used metrics \cite{han2022adbench}: (1) the Area Under the Receiver Operating Characteristic Curve (i.e., AUROC) and (2) the Area Under the Precision-Recall Curve (i.e., AUPRC). 
Both are the higher, the better.

\noindent
\textbf{LLMs and Hardware.} 
\add{We select three LLMs as main backbones: (1) Llama 3.1 8B Instruct (referred to as Llama 3.1)~\citep{llama}, (2) GPT-4o~\citep{gpt4o}, and (3) DeepSeek-V3~\citep{deepseekv3}.
Llama 3.1 represents an open-source model with accessible size and cost, GPT-4o serves as a closed-source model with advanced capabilities, and DeepSeek-v3 represents the Mixture-of-Experts (MoE) architecture. 
}

Llama 3.1 runs on one NVIDIA RTX 6000 Ada GPU with 48 GB RAM.
GPT-4o \add{and DeepSeek-V3} are accessed through official APIs.
Seed is set $=42$ for reproducibility.
Specific experimental settings are  highlighted separately in each subsequent task.

\def\doubleunderline#1{\underline{\underline{#1}}}
\begin{table*}[!t]
\vspace{-0.15in}
\setlength{\abovecaptionskip}{4pt}
\setlength{\belowcaptionskip}{4pt}
\caption{Performance comparison of LLM-based detectors and baseline methods across five datasets, evaluated under two settings as described in \S\ref{sec:ad_design} with AUROC and AUPRC as the metrics (higher (\raisebox{0.2ex}{$\uparrow$}), the better).
Complete results are provided in Appx.~\ref{tab:appx_pre_dataset}.
The \textbf{best} results are highlighted in bold, and the \underline{second-best} results are underlined. 
}

\label{tab:ad_results}
\centering
\setlength{\tabcolsep}{2.1pt}
\renewcommand{\arraystretch}{0.95}
\fontsize{8}{12}\selectfont 
\begin{tabular}{c cc cc cc cc cc}
\toprule
\multicolumn{1}{c}{\multirow{2}{*}{\textbf{Settings}}} & \multicolumn{2}{c}{\raisebox{0.3ex}{\textbf{AG News}}} & \multicolumn{2}{c}{\raisebox{0.3ex}{\textbf{BBC News}}} & \multicolumn{2}{c}{\raisebox{0.3ex}{\textbf{IMDB Reviews}}} & \multicolumn{2}{c}{\raisebox{0.3ex}{\textbf{N24 News}}} & \multicolumn{2}{c}{\raisebox{0.3ex}{\textbf{SMS Spam}}} \\ \cline{2-11}
\multicolumn{1}{c}{} & \raisebox{-0.3ex}{AUROC} \raisebox{-0.1ex}{$\uparrow$} & \raisebox{-0.3ex}{AUPRC} \raisebox{-0.1ex}{$\uparrow$} & \raisebox{-0.3ex}{AUROC} \raisebox{-0.1ex}{$\uparrow$} & \raisebox{-0.3ex}{AUPRC} \raisebox{-0.1ex}{$\uparrow$} & \raisebox{-0.3ex}{AUROC} \raisebox{-0.1ex}{$\uparrow$} & \raisebox{-0.3ex}{AUPRC} \raisebox{-0.1ex}{$\uparrow$} & \raisebox{-0.3ex}{AUROC} \raisebox{-0.1ex}{$\uparrow$} & \raisebox{-0.3ex}{AUPRC} \raisebox{-0.1ex}{$\uparrow$} & \raisebox{-0.3ex}{AUROC} \raisebox{-0.1ex}{$\uparrow$} & \raisebox{-0.3ex}{AUPRC} \raisebox{-0.1ex}{$\uparrow$} \\ \midrule
\multicolumn{1}{c}{} & \multicolumn{10}{c}{\raisebox{0.3ex}{\textbf{Llama 3.1 8B Instruct}}} \\
\multicolumn{1}{l}{(1) with $\mathcal{C}_{\text{normal}}$} & 0.8226 & 0.4036 & 0.7910 & 0.3602 & 0.7373 & 0.3474 & 0.6267 & 0.1130 & 0.7558 & 0.2884\\
\multicolumn{1}{l}{(2) with $\mathcal{C}_{\text{normal}}$, $\mathcal{C}_{\text{anomaly}}$} & 0.8754 & 0.3998 & 0.8612 & 0.3960 & {0.8625} & 0.4606 & {0.8784} & {0.3802} & {0.9487} & {0.6361} \\
\midrule
\multicolumn{1}{c}{} & \multicolumn{10}{c}{\raisebox{0.3ex}{\textbf{GPT-4o}}} \\
\multicolumn{1}{l}{(1) with $\mathcal{C}_{\text{normal}}$} & \textbf{0.9332} & \underline{0.7207} & {0.9574} & {0.8432} & {0.9349} & {0.7823} & 0.7674 & 0.3252 & 0.7940 & 0.5568\\
\multicolumn{1}{l}{(2) with $\mathcal{C}_{\text{normal}}$, $\mathcal{C}_{\text{anomaly}}$} & \underline{0.9293} & {0.6310} & \textbf{0.9919} & \textbf{0.9088} & \textbf{0.9668} & \underline{0.8465} & \textbf{0.9902} & \textbf{0.9009} &\textbf{0.9862} & \textbf{0.8953}\\
\midrule
\multicolumn{1}{c}{} & \multicolumn{10}{c}{\raisebox{0.3ex}{\textbf{DeepSeek-V3}}} \\
\multicolumn{1}{l}{(1) with $\mathcal{C}_{\text{normal}}$} & 0.9104 & 0.6442 & 0.8206 & 0.5604 & 0.8544 & 0.6808 & 0.8207 & 0.4495 & 0.8797 & 0.5963 \\
\multicolumn{1}{l}{(2) with $\mathcal{C}_{\text{normal}}$, $\mathcal{C}_{\text{anomaly}}$} & 0.9273 & \textbf{0.7817} & 0.9581 & \underline{0.8972} & \underline{0.9626} & \textbf{0.8569} & \underline{0.9514} & \underline{0.7730} & \underline{0.9535} & \underline{0.7914} \\
\midrule
\multicolumn{1}{c}{\multirow{2}{*}{\textbf{Best Baselines}}} & \multicolumn{2}{c}{{\rule{0pt}{10pt}\textbf{OpenAI + LUNAR}}} & \multicolumn{2}{c}{{\textbf{OpenAI + LUNAR}}} & \multicolumn{2}{c}{{\textbf{OpenAI + ECOD}}} & \multicolumn{2}{c}{{\textbf{OpenAI + LUNAR}}} & \multicolumn{2}{c}{{\textbf{DATE}}} \\
\multicolumn{1}{c}{} & 0.9226 & {0.6918} & \underline{0.9732} & {0.8653} & 0.7366 & 0.5165 & 0.8320 & {0.4425} & 0.9398 & 0.6112 \\\noalign{\vspace{2pt}}\cline{1-11}
\multicolumn{1}{c}{\multirow{2}{*}{\textbf{Second-best Baseline}}} & \multicolumn{2}{c}{{\rule{0pt}{11pt}\textbf{OpenAI + LOF}}} & \multicolumn{2}{c}{{\textbf{OpenAI + LOF}}} & \multicolumn{2}{c}{{\textbf{OpenAI + DeepSVDD}}} & \multicolumn{2}{c}{{\textbf{OpenAI + LOF}}} & \multicolumn{2}{c}{{\textbf{OpenAI + LOF}}} \\
\multicolumn{1}{c}{} & 0.8905 & 0.5443 & 0.9558 & 0.7714 & 0.6563 & 0.3278 & 0.7806 & 0.2248 & 0.7862 & 0.2450\\

\bottomrule
\end{tabular}
\vspace{-10pt}
\end{table*}

\vspace{-3ex}
\section{Task 1: LLM for Zero-shot Detection}
\label{sec:ad}
\vspace{-0.5ex}
\subsection{Motivation}
\vspace{-0.3ex}

Classical AD methods often require extensive training data—either labeled for supervised methods or unlabeled for unsupervised ones—which is time-consuming and costly \cite{han2022adbench}. 
In addition, setting up and tuning these models for real-world scenarios can be challenging and slow.

LLMs offer a practical alternative \cite{kaize_survey}. 
With their broad pre-trained knowledge, they can perform zero-shot detection without additional training data. 
Their ability to understand language context and semantics makes them suitable for recognizing anomalies by logical reasoning.
They can also explain their predictions, improving interpretability and trustworthiness \cite{huang2024position}, which is important in sensitive domains such as healthcare, finance, and cybersecurity.

\vspace{-0.5ex}
\subsection{Problem Statement and Designs}
\vspace{-0.3ex}
\label{sec:ad_design}

\begin{problem}[Zero-shot AD via LLMs]
Given a test set $\mathcal{D}_{\text{test}} = \left\{x_1, x_2, \dots, x_n\right\}$ of text samples, where each sample $x_i$ belongs to either a normal category or an anomaly category, the objective is to identify the anomalous samples using a pre-trained LLM $f_{\text{LLM}}$ in a zero-shot setting without any task-specific training data.
\end{problem}

\noindent
\textbf{Evaluation Protocol.}
We consider two settings, each reflecting different levels of prior knowledge:


\begin{itemize}[nosep, itemsep=3pt, leftmargin=*]
    \item \textbf{Normal Only:} We provide only the normal category name(s) $\mathcal{C}_{\text{normal}}$. This matches scenarios where normal behavior is known but anomalies are uncertain or emerging.
    \item \textbf{Normal + Anomaly:} We provide both normal and anomaly category names, $\mathcal{C}_{\text{normal}}$ and $\mathcal{C}_{\text{anomaly}}$. This setting reflects situations where some information on anomalies is available, helping the LLM reason about what is anomalous.
\end{itemize}

\noindent
The detection process is defined as:
\begin{equation}
    \begin{aligned}
        \label{eq:ad_prompt}
        \mathcal{P} &= T\left(x_i, \mathcal{C}_{\text{normal}}, \mathcal{C}_{\text{anomaly}}^*\right)\\
        \left(r, s\right) &= f_{\text{LLM}}\left(\mathcal{P}\right)
    \end{aligned}
\end{equation}
Here, $T(\cdot)$ constructs the prompt $\mathcal{P}$ for a test sample $x_i$, including known category information. 
The anomaly category is included only in the ``Normal + Anomaly'' setting, denoted as $\mathcal{C}_{\text{anomaly}}^*$. 
The LLM $f_{\text{LLM}}$ processes the prompt to produce a verbal anomaly score $s$ and an explanation $r$ that describes the reasoning. 
This setup allows a systematic evaluation of LLMs in zero-shot AD, using prompt-based inference to handle different levels of prior knowledge.
{See details in Appx.~\ref{appx:ad}.}

\vspace{-0.5ex}
\subsection{Results, Insights, and Future Directions}
\vspace{-0.3ex}
\label{sec:ad_results}
We select Llama 3.1, GPT-4o, \add{and DeepSeek-V3} as zero-shot detectors. Temperature is set as $=0$ for stable outputs. 



\noindent
\textbf{\textit{LLMs are effective in zero-shot AD, surpassing existing training-based AD algorithms}.}  
We compare LLM-based zero-shot detectors with top baselines across five datasets in Table~\ref{tab:ad_results}. GPT-4o \add{and DeepSeek-V3} consistently outperform baselines; Llama 3.1 shows competitive performance when anomaly information is available.
Despite operating with limited prior information, LLMs exhibit significant potential for anomaly detection tasks.
These results highlight the strength of LLMs in zero-shot AD scenarios.


\noindent\textbf{\textit{Additional context helps.}} 
Table~\ref{tab:ad_results} shows that LLM-based detectors achieve improved AUROC and AUPRC when transitioning from setting ``Normal Only'', which uses only $\mathcal{C}_{\text{normal}}$, to setting ``Normal + Anomaly'', which includes both $\mathcal{C}_{\text{normal}}$ and $\mathcal{C}_{\text{anomaly}}$. These results indicate that richer contextual information improves the LLMs’ ability to distinguish anomalous samples and enhances detection performance.

\noindent
\textbf{\textit{Future Direction 1: Improve Context Integration.}}  
Providing additional context improves detection, as seen in ``with $\mathcal{C}_{\text{normal}}$, $\mathcal{C}_{\text{anomaly}}$.'' Future work may involve more systematic ways to integrate domain-specific details, 
such as prompt design or retrieval-augmented methods \cite{ad_rag}.

\noindent
\textbf{\textit{Future Direction 2: Optimize for Real-world Deployment.}} 
Despite their effectiveness, LLM-based zero-shot AD is inherently time-consuming and costly \add{during the inference} \cite{llm_cost}.
Reducing computational overhead is important for deploying LLMs in real settings, especially for AD applications, which are often time-critical. 
Methods like quantization \cite{ad_q_lora, ad_q_smoothquant}, pruning \cite{ad_pruning_1, ad_pruning_fuamoeballm}, and knowledge distillation \cite{ad_kd_1, ad_kd_2} can help reduce the model size and inference time while maintaining good performance.


\noindent
\section{Task 2: LLM for Data Augmentation}
\label{sec:aug}
\subsection{Motivation}

Data augmentation (DA) in AD aims to produce additional samples to improve model training under data scarcity \cite{yoo2023dsv}.
However, traditional methods often struggle to capture the complexity of natural language, potentially causing a shift in domain characteristics \cite{da_disadvantage}.
LLMs offer a solution, using their broad pre-trained knowledge and autoregressive learning objectives to generate contextually relevant data with better semantic understanding \cite{kaize_survey}.

In addition, LLMs can generate textual descriptions~\cite{kaize_survey} that assist the LLM-based detectors in \S\ref{sec:ad}. 
For example, by producing descriptions of \textbf{known categories}, LLMs help detectors establish distant associations between normal and anomalous samples \cite{aug_desc_22, aug_desc_24}.


Thus, We examine two approaches that address data scarcity and improve semantic reasoning:
\begin{enumerate}[nosep, itemsep=3pt, leftmargin=*]
    \item 
    (\S \ref{subsec:aug_syn}) generates \textit{synthetic samples} to improve \underline{training-based AD models}.
    \item 
    (\S \ref{subsec:aug_desc}) produces \textit{category descriptions} to refine prompts and enhance \underline{LLM-based detectors}.
\end{enumerate}

\subsection{Generating Synthetic Samples for Training-based AD Models}
\label{subsec:aug_syn}

\begin{problem}[Synthetic DA via LLMs]
Given a small training set $\mathcal{D}_{\text{small\_train}} = \left\{x_1, x_2, \dots, x_m\right\}$ of normal samples, the goal is to produce a synthetic dataset $\mathcal{D}_{\text{synth}} = \left\{\tilde{x}_1, \tilde{x}_2, \dots, \tilde{x}_n\right\}$ using a pre-trained LLM $f_{\text{LLM}}$. The combined dataset $\mathcal{D}_{\text{DA}} = \mathcal{D}_{\text{small\_train}} \cup \mathcal{D}_{\text{synth}}$ is used to train an unsupervised AD method $M$, improving performance compared to using $\mathcal{D}_{\text{small\_train}}$ alone.
\end{problem}

\noindent
\textbf{Evaluation Protocol.}
To evaluate the impact of LLM-generated synthetic data, we set unsupervised AD baselines listed in Appx.~\ref{appx:pre_baseline} in a scenario with limited training data. 
LLMs are then utilized to generate a synthetic training dataset. However, direct prompting often leads to highly repetitive outputs, even with high decoding temperatures \cite{aug_synth_multi_step}. Additionally, LLMs face constraints such as token limits and challenges in processing long contexts \cite{llm_long_context}.
To address these issues, we adopt a \textbf{multi-step} strategy:

\label{sec:aug_syn_overview}
\begin{itemize}[nosep, itemsep=3pt, leftmargin=*]
    \item \textit{Step1: Keyword Generation:} Generate groups of keywords in one inquiry. Each group contains three keywords with a different level of granularity: broad/general, intermediate, or fine-grained.
    \item \textit{Step2: Sample Generation:} For each keyword group, generate one synthetic sample $\tilde{x}_i$.
\end{itemize}

Separating keyword generation from sample creation and enforcing different granularity levels ensures controlled variability and prevents overly long or repetitive outputs. This results in more contextually rich and diverse synthetic samples.

\add{To scale up further, we generate synthetic data in multiple rounds. In each round, we adjust the random seed, decoding temperature, and prompt template to ensure diversity.}
Further details are provided in Appx.~\ref{appx:aug_syn}

\begin{figure}[!t]
\setlength{\abovecaptionskip}{3pt}
\includegraphics[width=0.48\textwidth]{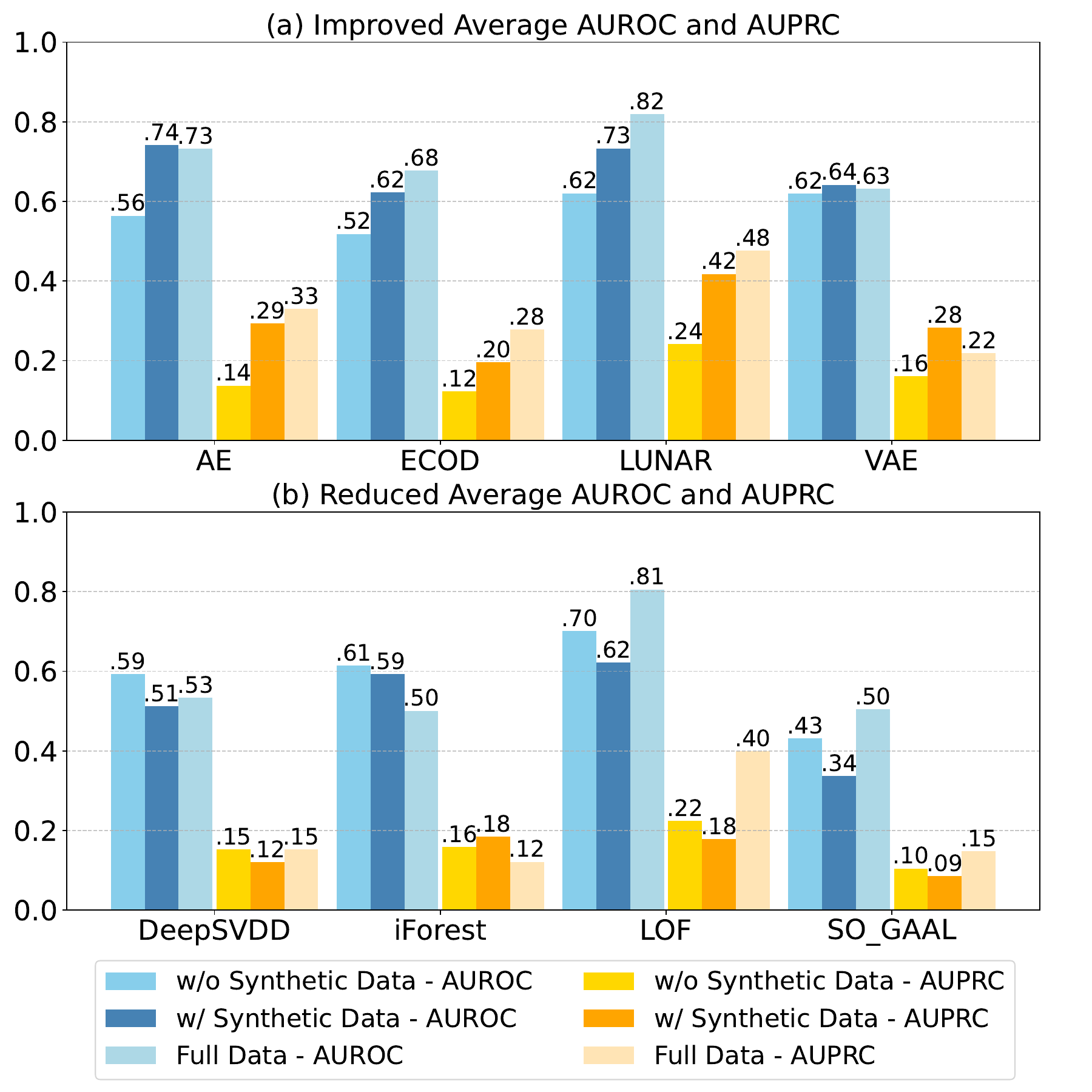}
   \caption{Average performance over five datasets of AD baselines trained on limited data, w/ or w/o LLM-generated synthetic data, and on full datasets across five datasets. (a) Detectors that benefit from augmentation. (b) Detectors that degrade with augmentation.}
   \label{fig:aug_synth}
   \vspace{-16pt}
\end{figure}

\def\doubleunderline#1{\underline{\underline{#1}}}
\begin{table*}[t]
\vspace{-0.15in}
\setlength{\abovecaptionskip}{4pt}
\caption{Performance (and $\triangle$ changes) of LLM-based detectors \textbf{with augmented descriptions} under two settings in \S\ref{sec:ad_design}. The description generators and LLM-based detectors adopt the same backbone.
Values in brackets indicate changes compared to the results in Table \ref{tab:ad_results}. \textcolor{ForestGreen}{Green} denotes for improvements and \textcolor{BrickRed}{red} for declines. 
\textbf{Changes below $0.03$ are not colored} for better visualization, also reflecting minor fluctuations. 
}

\label{tab:aug_desc_results}
\centering
\setlength{\tabcolsep}{2.1pt}
\renewcommand{\arraystretch}{0.95}
\fontsize{8}{12}\selectfont 
\begin{tabular}{c cc cc cc cc cc}
\toprule
\multicolumn{1}{c}{\multirow{2}{*}{\textbf{Settings}}} & \multicolumn{2}{c}{\raisebox{0.3ex}{\textbf{AG News}}} & \multicolumn{2}{c}{\raisebox{0.3ex}{\textbf{BBC News}}} & \multicolumn{2}{c}{\raisebox{0.3ex}{\textbf{IMDB Reviews}}} & \multicolumn{2}{c}{\raisebox{0.3ex}{\textbf{N24 News}}} & \multicolumn{2}{c}{\raisebox{0.3ex}{\textbf{SMS Spam}}} \\ \cline{2-11}
\multicolumn{1}{c}{} & \raisebox{-0.3ex}{AUROC} \raisebox{-0.1ex}{$\uparrow$} & \raisebox{-0.3ex}{AUPRC} \raisebox{-0.1ex}{$\uparrow$} & \raisebox{-0.3ex}{AUROC} \raisebox{-0.1ex}{$\uparrow$} & \raisebox{-0.3ex}{AUPRC} \raisebox{-0.1ex}{$\uparrow$} & \raisebox{-0.3ex}{AUROC} \raisebox{-0.1ex}{$\uparrow$} & \raisebox{-0.3ex}{AUPRC} \raisebox{-0.1ex}{$\uparrow$} & \raisebox{-0.3ex}{AUROC} \raisebox{-0.1ex}{$\uparrow$} & \raisebox{-0.3ex}{AUPRC} \raisebox{-0.1ex}{$\uparrow$} & \raisebox{-0.3ex}{AUROC} \raisebox{-0.1ex}{$\uparrow$} & \raisebox{-0.3ex}{AUPRC} \raisebox{-0.1ex}{$\uparrow$} \\ \midrule
\multicolumn{1}{c}{} & \multicolumn{10}{c}{\raisebox{0.3ex}{\textbf{Llama 3.1 8B Instruct}}} \\
\multicolumn{1}{l}{\multirow{2}{*}{(1) with $\mathcal{C}_{\text{normal}}$}} & 0.8081& 0.3588 & 0.7802 & 0.3006 & 0.9039 & 0.6272 & 0.6651 & 0.1383 & 0.7456 & 0.2225 \\
\multicolumn{1}{c}{} & {(-0.0145)} & \textcolor{BrickRed}{(-0.0448)} & {(-0.0108)} & \textcolor{BrickRed}{(-0.0596)} & \textcolor{ForestGreen}{(+0.1666)} & \textcolor{ForestGreen}{(+0.2798)} & \textcolor{ForestGreen}{(+0.0384)} & {(+0.0253)} & {(-0.0102)} & \textcolor{BrickRed}{(-0.0659)} \\
\vspace{-10pt}\\
\cline{2-11}
\vspace{-10pt}\\
\multicolumn{1}{l}{\multirow{2}{*}{(2) with $\mathcal{C}_{\text{normal}}$, $\mathcal{C}_{\text{anomaly}}$}} & 0.9046 & 0.5097 & 0.9089 & 0.6531 & 0.9351 & 0.6369 & 0.7900 & 0.2396 & 0.9413 & 0.7018 \\
\multicolumn{1}{c}{} & {(+0.0292)} & \textcolor{ForestGreen}{(+0.1099)} & \textcolor{ForestGreen}{(+0.0477)} & \textcolor{ForestGreen}{(+0.2571)} & \textcolor{ForestGreen}{(+0.0726)} & \textcolor{ForestGreen}{(+0.1763)} & \textcolor{BrickRed}{(-0.0884)} & \textcolor{BrickRed}{(-0.1406)} & {(-0.0074)} & \textcolor{ForestGreen}{(+0.0657)} \\
\midrule
\multicolumn{1}{c}{} & \multicolumn{10}{c}{\raisebox{0.3ex}{\textbf{GPT-4o}}} \\
\multicolumn{1}{l}{\multirow{2}{*}{(1) with $\mathcal{C}_{\text{normal}}$}} & 0.9255 & 0.6985 & 0.9611 & 0.8162 & 0.9572 & 0.8307 & 0.8792 & 0.5399 & 0.8365 & 0.4765 \\
\multicolumn{1}{c}{} & {(-0.0077)} & {(-0.0222)} & {(+0.0037)} & {(-0.0270)} & {(+0.0223)} & \textcolor{ForestGreen}{(+0.0484)} & \textcolor{ForestGreen}{(+0.1118)} & \textcolor{ForestGreen}{(+0.2147)} & \textcolor{ForestGreen}{(+0.0425)} & \textcolor{BrickRed}{(-0.0803)} \\  
\vspace{-10pt}\\
\cline{2-11}
\vspace{-10pt}\\
\multicolumn{1}{l}{\multirow{2}{*}{(2) with $\mathcal{C}_{\text{normal}}$, $\mathcal{C}_{\text{anomaly}}$}} & 0.9331 & 0.6659 & 0.9849 & 0.8998 & 0.9855 & 0.9219 & 0.9895 & 0.8680 & 0.9800 & 0.8889 \\
\multicolumn{1}{c}{} & \textcolor{ForestGreen}{(+0.0038)} & \textcolor{ForestGreen}{(+0.0349)} & {(-0.0070)} & {(-0.0090)} & {(+0.0187)} & \textcolor{ForestGreen}{(+0.0754)} & {(-0.0007)} & \textcolor{BrickRed}{(-0.0329)} & {(-0.0062)} & {(-0.0064)} \\
\midrule
\multicolumn{1}{c}{} & \multicolumn{10}{c}{\raisebox{0.3ex}{\textbf{DeepSeek-V3}}} \\
\multicolumn{1}{l}{\multirow{2}{*}{(1) with $\mathcal{C}_{\text{normal}}$}} & 0.8791 & 0.5180 & 0.8800 & 0.6170 & 0.9612 & 0.7888 & 0.8261 & 0.3949 & 0.9262 & 0.6128 \\
\multicolumn{1}{c}{} & \textcolor{BrickRed}{(-0.0482)} & \textcolor{BrickRed}{(-0.1262)} & \textcolor{ForestGreen}{(+0.0594)} & \textcolor{ForestGreen}{(+0.0566)} & \textcolor{ForestGreen}{(+0.1068)} & \textcolor{ForestGreen}{(+0.1080)} & {(+0.0054)} & \textcolor{BrickRed}{(-0.0546)} & \textcolor{ForestGreen}{(+0.0465)} & {(+0.0165)} \\  
\vspace{-10pt}\\
\cline{2-11}
\vspace{-10pt}\\
\multicolumn{1}{l}{\multirow{2}{*}{(2) with $\mathcal{C}_{\text{normal}}$, $\mathcal{C}_{\text{anomaly}}$}} & 0.9231 & 0.6492 & 0.9577 & 0.9106 & 0.9793 & 0.9241 & 0.9591 & 0.8072 & 0.9522 & 0.8697 \\
\multicolumn{1}{c}{} & {(-0.0042)} & \textcolor{BrickRed}{(-0.1325)} & {(-0.0004)} & {(+0.0134)} & {(+0.0167)} & \textcolor{ForestGreen}{(+0.0672)} & {(+0.0083)} & \textcolor{ForestGreen}{(+0.0342)} & {(-0.0013)} & \textcolor{ForestGreen}{(+0.0783)} \\
\bottomrule
\end{tabular}
\vspace{-10pt}
\end{table*}

\noindent
\textbf{Results, Insights, and Future Directions.}  
We use GPT-4o with temperature varying from 0.7 to 1.0 in multi-round synthetic generation. Table~\ref{tab:aug_synth_results} presents the complete results.

\noindent
\textbf{\textit{LLM-generated synthetic data effectively improves AD performance.}}
\add{Our results show that LLM-generated synthetic data significantly enhances AD performance for several detectors. As illustrated in Figure~\ref{fig:aug_synth}(a), models like AE, ECOD, LUNAR, and VAE achieve substantial AUROC and AUPRC improvements when synthetic samples are included alongside limited real data. Notably, these models often close the gap between limited-data performance and full-data performance, demonstrating that synthetic generation can effectively compensate for data scarcity. }



\add{\noindent
\textbf{\textit{Performance impact varies across models.}}
The effectiveness of synthetic generation is not consistent across all models. Methods relying on fixed geometric assumptions—such as DeepSVDD, iForest, and LOF—often degrade after augmentation (Figure~\ref{fig:aug_synth}(b)). The variance introduced by synthetic data may expand DeepSVDD’s hypersphere, perturb iForest’s isolation statistics, or blur LOF’s local-density estimates, weakening the separation between normal and anomalous points. Similarly, SO\_GAAL’s adversarial training objective may become unstable as the variance may widen the definition of normal data, complicating discriminator convergence. In contrast, models like AE, ECOD, LUNAR, and VAE substantially benefit from synthetic data. Their reconstruction (AE, VAE), empirical-distribution (ECOD), or graph-aggregation (LUNAR) objectives may leverage the enriched embedding manifold, leading to more robust representations and improved detection performance. In short, synthetic generation effectively enhances detectors that learn flexible representations but can impair those reliant on fixed geometric criteria or unstable adversarial objectives.}

\noindent
\textbf{\textit{Future Direction 3: Balance Diversity and Alignment in Synthetic Data.}}  
Future work should investigate techniques to balance the diversity of synthetic samples with their semantic alignment to real-world distributions.
Excessive diversity risks producing samples that deviate too far from the target domain, while insufficient diversity may fail to address data scarcity and limit generalization \cite{synth_challenges}. 
Potential strategies include adjusting the prompt engineering process, using retrieval-augmented LLMs, embedding-based filters to steer generation \cite{synth_steer}, and incorporating human-in-the-loop interventions \cite{synth_increasing} to refine synthetic data quality and improve downstream AD performance.




\vspace{-0.3ex} 
\subsection{Generating Category Descriptions for LLM-based Detectors}
\label{subsec:aug_desc}
\begin{problem}[Description DA via LLMs]
Given category names $\mathcal{C}_{\text{normal}}$ and, optionally, $\mathcal{C}_{\text{anomaly}}$, the objective is to generate comprehensive textual descriptions $d_{\text{normal}}$ and $d_{\text{anomaly}}$ using a pre-trained LLM $f_{\text{LLM}}$. 
These descriptions are then incorporated into the prompts of LLM-based detectors, aiming to improve their performance compared to using category names alone.
\end{problem}
\noindent
\textbf{Evaluation Protocol.}
Extending the zero-shot detection from \S\ref{sec:ad}, we employ LLMs to produce category descriptions that offer richer semantic signals beyond simple category names. 
Specifically, for each normal and anomaly category, we generate $d_{\text{normal}}$ and $d_{\text{anomaly}}$ based on the category names and the dataset’s context. These descriptions can highlight distinctive features, typical lexical patterns, or behavioral characteristics that define normal or anomalous classes. 
By incorporating these descriptions into the prompt, we update Eq.~(\ref{eq:ad_prompt}) as:
\begin{equation}
    \begin{aligned}
        \mathcal{P} = T\biggl(x_i,& \left(\mathcal{C}_{\text{normal}}, \fcolorbox{blue}{white}{$d_{\text{normal}}$}\right),\\
        &\left(\mathcal{C}_{\text{anomaly}}, \fcolorbox{blue}{white}{$d_{\text{anomaly}}$}\right)^*\biggr)
    \end{aligned}
\end{equation}
where $(\mathcal{C}_{\text{anomaly}}, d_{\text{anomaly}})^*$ applies only in the ``Normal + Anomaly'' setting (see \S\ref{sec:ad_design}). 
By enriching category names with descriptions (highlighted with blue boxes), we enhance the LLM’s ability to reason about subtle category distinctions. More details are provided in Appx.~\ref{appx:aug_desc}.


\noindent
\textbf{Results, Insights, and Future Directions.}
We utilize Llama 3.1, GPT-4o, \add{and DeepSeek-V3} to generate category descriptions. We set the temperature $=0.5$ to balance the diversity and precision.

\noindent  
\textbf{\textit{Augmented descriptions improve LLM-based AD}.}  
As shown in Table~\ref{tab:aug_desc_results}, incorporating category descriptions increases performance in most datasets. 
This suggests that the added semantic information helps LLM-based detectors discriminate anomalous samples more effectively.
For example, in the ``IMDB Reviews'' dataset, providing richer textual representations of classes translates to noticeable gains in both metrics across LLMs.



\noindent
\textbf{\textit{Future Direction 4: Select Representative Samples.}}  
An effective way to refine enhanced information is to ground it in representative samples from the dataset.
Sampling strategies based on clustering \cite{desc_cluster} or diversity maximization \cite{desc_diverse} can identify prototype examples that guide LLMs to produce more tailored and context-aware descriptions. 
By referencing these representative samples, future methods may generate more refined information that better distinguishes between normal data and anomalies, ultimately improving AD performance.

\begin{figure}[!t]
\includegraphics[width=0.48\textwidth]{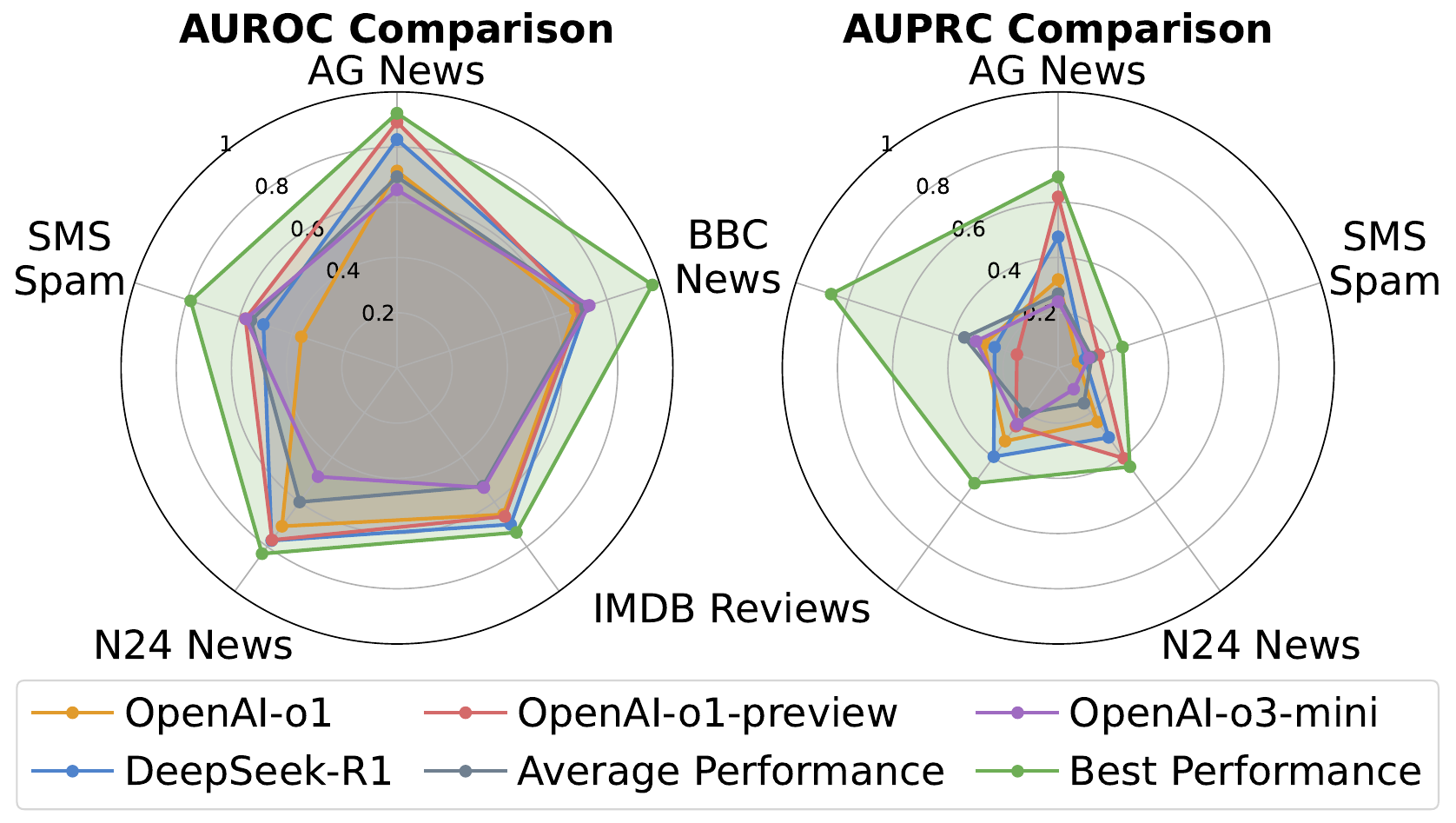}
    \caption{Model selection results across five datasets. We display the average AUROC and AUPRC of models recommended by querying each reasoning LLM five times (duplicates allowed). ``Best Performance'' marks the highest performance achieved by any baseline model for each dataset, while ``Average Performance'' denotes the mean performance across all baseline models.
   }
   \label{fig:ums_results}
   \vspace{-5pt}
\end{figure}

\section{Task 3: LLM for AD Model Selection}
\label{sec:ums}
\subsection{Motivation}
Unsupervised model selection (UMS) is critical for identifying the most suitable AD model by aligning its features with the attributes of a given dataset and the task's requirements. 
Given the diverse range of AD models available and the absence of a universal solution, effective UMS is essential to ensure optimal performance. 
Traditional UMS methods often rely on historical performance data or domain-specific expertise; however, such data may be unavailable or irrelevant for novel or evolving datasets \cite{zhao2021automatic,zhao2024towards}.

Inspired by recent research~\citep{qin2024metaood, chen2024pyod, wei2025tsselect}, LLMs offer a promising zero-shot alternative by utilizing their extensive pre-trained knowledge to analyze datasets and recommend suitable models without relying on past performance metrics. 
They can streamline the model selection process, reducing manual overhead and domain knowledge requirements while also improving adaptability to novel data scenarios.

\subsection{Problem Statement and Designs}
\label{subsec:ums_design}
\begin{problem}[Zero-shot UMS via LLMs]
Given a dataset $\mathcal{D} = \left\{x_1, x_2, \dots, x_n\right\}$ and a set of AD models $\mathcal{M} = \left\{M_1, M_2, \dots, M_m\right\}$, the task is to identify a suitable model $M^* \in \mathcal{M}$ using a pre-trained LLM $f_{\text{LLM}}$, based solely on provided information about the dataset and the candidate models.
\end{problem}

\noindent
\textbf{Evaluation Protocol.}
To enable LLM-based zero-shot UMS, we provide structured, detailed information of both the dataset and the candidate models:

\begin{itemize}[nosep, itemsep=3pt, leftmargin=*]
    \item \textbf{Dataset Description:} dataset name, size, background, normal and anomaly categories, text-length statistics (average, maximum, minimum, and standard deviation), and representative samples of both normal and anomalous data. These attributes help the LLM understand the dataset's structure, complexity, and potential challenges, and are generally easy to obtain for new datasets.
    \item \textbf{Model Description:} abstracts from published AD papers describing each candidate model. These abstracts highlight key model features, underlying assumptions, and targeted use cases. 
    By examining these summaries, the LLM can align dataset attributes with model strengths, improving the relevance of its recommendations.
\end{itemize}

We then construct prompts that combine these datasets and model descriptions, asking the LLM to select and justify a recommended model. Further details about the prompt format and implementation can be found in Appx.~\ref{appx:ums}.

\def\doubleunderline#1{\underline{\underline{#1}}}
\begin{table}[!t]
\setlength{\abovecaptionskip}{6pt}
\setlength{\belowcaptionskip}{6pt}
\caption{
The top 2 frequent picks made by each LLM.
Counts are aggregated over 25 queries (5 per dataset).
}

\label{tab:ums_top_count}
\centering
\setlength{\tabcolsep}{2pt}
\renewcommand{\arraystretch}{0.95}
\fontsize{7.8}{12}\selectfont 
\begin{tabular}{l l}
\toprule
\textbf{LLMs} & \textbf{Top-2 Picks (counts)} \\
\midrule
OpenAI-o1-preview\  & OpenAI+LUNAR (13), OpenAI+ECOD (8) \\
DeepSeek-R1 & OpenAI+ECOD (16), OpenAI+LUNAR (6) \\
OpenAI-o1 & OpenAI+DeepSVDD (11), OpenAI+iForest (7) \\
OpenAI-o3-mini & BERT+DeepSVDD (10), OpenAI+ECOD (6) \\
\bottomrule
\vspace{-15pt}
\end{tabular}
\end{table}

\subsection{Results, Insights, and Future Directions}
\label{sec:ums_results}
The UMS scenario requires sophisticated reasoning. \add{We select recent enhanced reasoning models, including OpenAI-o1-preview and OpenAI-o1~\citep{gpto1}, OpenAI-o3-mini~\citep{gpto3-mini}, and DeepSeek-R1~\citep{deepseek-r1}}.

\noindent
\textbf{\textit{LLM recommendations demonstrate strong potential.}}
Figure~\ref{fig:ums_results} presents the model selection performance of \add{four reasoning LLMs} across five datasets, compared against two reference baselines: (\textit{i}) the \textbf{best} result achieved by any baseline model, representing the performance upper bound; and (\textit{ii}) the \textbf{average} performance of all baseline models, reflecting random model selection. In most cases, the AD performance of LLM-recommended models surpasses the average baseline and even approaches the best-performing model. These results highlight the strong potential of LLM-based reasoning to identify effective AD models using only public information, without reliance on historical performance or domain specialists.

\add{\noindent
\textbf{\textit{LLMs exhibit inherent yet context-sensitive selection biases.}}
The aggregated selection results in Table~\ref{tab:ums_top_count} highlight distinct model-selection preferences among the four LLMs. For example, OpenAI-o1-preview often recommends OpenAI+LUNAR, a consistently strong model in our benchmarks. In contrast, OpenAI-o3-mini prefers Bert+DeepSVDD, a generally weaker option. To investigate whether these biases arise from internal priors or context-specific information, we conducted a selection experiment without any dataset or model details (Table~\ref{tab:ums_zero_shot}). Although each LLM still favors a distinct, limited set of models even without context, their specific preferences notably shift when introducing the context. This indicates that while LLMs possess intrinsic biases from their pretraining or tuning phases, their selections are also influenced by provided context information.}

\add{\noindent
\textbf{\textit{Context information improves selections, but justifications remain generic.}} When comparing model selection results with and without context (Table~\ref{tab:ums_top_count} - Table~\ref{tab:ums_zero_shot}), we notice a clear shift in model recommendations that generally better align with benchmark results.} Despite this improved selection accuracy, its explanations often remain generic and do not clearly link model selection to specific dataset characteristics. 
For example, in the ``AG News'' dataset, the OpenAI-o1-preview alternated between recommending ``OpenAI + LUNAR'' and ``OpenAI + ECOD,'' justifying choices with broad statements like ``effective for high-dimensional data'' or ``parameter-free scalability.'' Such non-specific rationales diminish interpretability and user trust, especially when understanding the rationale behind model choice is important.



\def\doubleunderline#1{\underline{\underline{#1}}}
\begin{table}[!t]
\vspace{-3pt}
\setlength{\abovecaptionskip}{5pt}
\setlength{\belowcaptionskip}{5pt}
\caption{
Selections by each LLM without any dataset or model context. Each LLM was queried five times.
}

\label{tab:ums_zero_shot}
\centering
\setlength{\tabcolsep}{2.1pt}
\renewcommand{\arraystretch}{0.95}
\fontsize{8}{12}\selectfont 
\begin{tabular}{l l}
\toprule
\textbf{LLMs} & \textbf{Context-Free Picks (counts)} \\
\midrule
OpenAI-o1-preview\  & OpenAI+LUNAR (3), OpenAI+VAE (2) \\
DeepSeek-R1 & OpenAI+LUNAR (3), OpenAI+ECOD (2) \\
OpenAI-o1 & OpenAI+VAE (4), OpenAI+LUNAR (1) \\
OpenAI-o3-mini & OpenAI+VAE (4), OpenAI+DeepSVDD (1) \\
\bottomrule
\vspace{-25pt}
\end{tabular}
\end{table}


\noindent
\textbf{\textit{Future Direction 5: Refine Input Specificity and Alleviate Biases.}}  
Future work should explore how to provide more dataset-specific details and mitigate potential LLM biases. 
Ambiguous or incomplete input information may cause the LLM to favor well-known models or those frequently encountered during training.
Ensuring detailed and balanced inputs, and exploring how inherent biases in LLMs affect recommendations, will be important steps to improve the fairness and reliability of LLM-based UMS \cite{bias}.

\noindent
\textbf{\textit{Future Direction 6: Enhancing Interpretability.}}  
Improving LLMs' capacity to produce transparent, dataset-tailored justifications for model selection decisions is key \cite{trust_llm}. 
Techniques such as fine-tuning with richly annotated explanations or using prompt engineering to explicitly request structured reasoning can encourage the LLM to articulate clear, context-sensitive arguments.
\section{Conclusion}
In this work, we presented \ours, the first comprehensive benchmark that integrates LLMs into three core aspects of anomaly detection in NLP: detection, data augmentation, and model selection.

Our results show that LLMs exhibit promising capabilities in zero-shot AD without task-specific training. LLM-generated synthetic data significantly boosted performance for models that learn flexible representations, while it may negatively impact models that rely on rigid geometric assumptions. Additionally, LLM-driven model selection frequently exceeded baseline performance, though explanations for these selections often lacked dataset-specific detail.

\section*{Future Directions}
Future research should focus on improving contextual prompts to enhance zero-shot AD capabilities while considering the cost, developing methods to balance diversity and domain alignment in synthetic data generation, and increasing the specificity and interpretability of LLM-generated model selection justifications. Expanding the \ours benchmark to include additional tasks and applications in different fields \cite{huang2024position,li2024political} also represents a valuable direction for broadening its impact. 
\section*{Broader Impact Statement}
\ours explores the use of LLMs in enhancing AD through zero-shot detection, data augmentation, and model selection. 
These contributions have the potential to significantly improve real-world AD systems in critical areas such as healthcare, finance, and cybersecurity. 
By enabling robust, adaptable, and efficient solutions for AD tasks, this research empowers practitioners to deploy systems responsive to novel challenges while reducing reliance on labeled data and extensive domain expertise.

\section*{Ethics Statement}
This study adheres to ethical guidelines, emphasizing considerations around fairness, transparency, and privacy in developing and applying LLM-based AD systems. 
We emphasize the importance of evaluating and mitigating biases in LLM recommendations, ensuring that outputs are equitable and unbiased. 
Moreover, privacy is preserved by relying on public data and avoiding the collection of sensitive information. 
Also, note that we used ChatGPT exclusively to improve minor grammar in the final manuscript text.


\section*{Limitations}
Despite promising results, several limitations remain. 
First, our evaluation is constrained to a narrow set of datasets with clear normal-anomaly distinctions, and our settings in AD and category descriptions in DA follow the structure of these datasets, limiting applicability to various domains with ambiguous anomaly definitions.
Second, UMS depends on simplistic input data and matching mechanisms. Furthermore, biases in LLM recommendations, such as favoring well-documented or familiar models, need further investigation.
Additionally, we do not explore few-shot learning or fine-tuning, which are widely adopted techniques for enhancing LLM performance and could offer valuable complementary insights for AD tasks.

\section*{Acknowledgments}

This work was partially supported by the National Science Foundation under Award No.~2428039 and No.~2346158.  
We also acknowledge the use of computational resources provided by the Advanced Cyberinfrastructure Coordination Ecosystem \cite{boerner2023access}: Services \& Support (ACCESS) program, supported by NSF grants \#2138259, \#2138286, \#2138307, \#2137603, and \#2138296. Specifically, this work used NCSA Delta GPU at the National Center for Supercomputing Applications (NCSA) through allocation CIS250073.  
Any opinions, findings, conclusions, or recommendations expressed in this material are those of the authors and do not necessarily reflect the views of the National Science Foundation.

\vspace{20pt}

\bibliography{custom,yue}

\clearpage
\newpage

\appendix
\label{sec:appendix}
\section*{Supplementary Material for \ours}
\setcounter{section}{0}
\setcounter{figure}{0}
\setcounter{table}{0}
\makeatletter 
\renewcommand{\thesection}{\Alph{section}}
\renewcommand{\theHsection}{\Alph{section}}
\renewcommand{\thefigure}{A\arabic{figure}} 
\renewcommand{\theHfigure}{A\arabic{figure}} 
\renewcommand{\thetable}{A\arabic{table}}
\renewcommand{\theHtable}{A\arabic{table}}
\makeatother

\renewcommand{\thetable}{A\arabic{table}}
\setcounter{equation}{0}
\renewcommand{\theequation}{A\arabic{equation}}

\section{Additional Details for Preliminaries}
\label{appx:pre}
\subsection{Datasets Details}
\label{appx:pre_dataset}
As briefly discussed in \S\ref{sec:pre_data_baseline}, we select five NLD AD datasets with high quality and a proper size sourced from \cite{nlp_adbench}: AG News, BBC News, IMDB Reviews, N24News, SMS Spam. These datasets are originally intended for NLP classification tasks and contain text samples categorized into multiple groups, with one designated anomalous. The training data comprises only normal samples. Table~\ref{appx:pre_dataset} provides a summary of dataset attributes, and Table~\ref{tab:appx_data_stat} presents the statistics of datasets that will be utilized in our tasks.

\begin{table}[h]
\setlength{\abovecaptionskip}{5pt}
\centering
\setlength{\tabcolsep}{8pt}
\renewcommand{\arraystretch}{0.9}
\fontsize{8}{13.5}\selectfont 
\begin{tabular}{l c c c c}
\toprule
\textbf{Dataset} & \textbf{Avg.} &\textbf{Max.} &\textbf{Min.} &\textbf{Std.} \\
\midrule
\textbf{AG News} & 190.1 & 959 & 35 & 61.7 \\
\textbf{BBC News} & 2,293.5 & 25,367 & 685 & 1,506.4 \\
\textbf{IMDB Reviews} & 1,289.2 & 12,498 & 65 & 980.5 \\
\textbf{N24 News} & 4,633.3 & 28,616 & 4 & 3,069.5 \\
\textbf{SMS Spam} & 78.7 & 790 & 4 & 60.8 \\
\bottomrule
\end{tabular}
\caption{Statistics of datasets including average, maximum, minimum, standard deviation of text length.}
\label{tab:appx_data_stat}
\vspace{-12pt}
\end{table}
\vspace{-0.8ex}
\subsection{Traditional Baselines Details}
\vspace{-0.5ex}
\label{appx:pre_baseline}
This study utilizes 18 traditional methods as baselines. We compare the performance of LLM-based anomaly detection methods with these baselines in \S\ref{sec:ad} and further enhance the baselines with LLM-generated synthetic data, demonstrating the effectiveness of augmentation in \S\ref{sec:aug_syn_overview}.

These methods are categorized into two groups:
\begin{itemize}[nosep, itemsep=3pt, leftmargin=*]
\item \textbf{End-to-end Methods}. These methods directly process raw text data to generate AD results:
    \begin{itemize}[nosep, itemsep=3pt, leftmargin=*]
    \item \textbf{CVDD}: Context Vector Data Description \cite{cvdd}. CVDD uses embeddings and self-attention to learn context vectors, detecting anomalies via deviations.
    \item \textbf{DATE}: Detecting Anomalies in Text via Self-Supervision of Transformers \cite{date}. DATE trains self-supervised transformers to identify anomalies in text.
    \end{itemize}
\item \textbf{Two-Step Methods}. These approaches first generate text embeddings using BERT~\cite{bert} or OpenAI's \textit{text-embedding-3-large}~\cite{openai2024embedding} and then apply traditional AD techniques to the embeddings.
    \begin{itemize}[nosep, itemsep=1pt, leftmargin=*]
    \item \textbf{AE}: AutoEncoder \cite{ae}. AE uses high reconstruction errors to detect anomalies.
    \item \textbf{DeepSVDD}: Deep Support Vector Data Description \cite{deepsvdd}. DeepSVDD identifies anomalies outside a hypersphere that encloses normal data representations.
    \item \textbf{ECOD}: Empirical-Cumulative-distribution-based Outlier Detection \cite{ecod}. ECOD flags point in distribution tails using empirical cumulative distributions.
    \item \textbf{IForest}: Isolation Forest \cite{iforest}. IForest isolates anomalies with fewer splits in random feature-based partitions.
    \item \textbf{LOF}: Local Outlier Factor \cite{lof}. LOF detects anomalies by comparing the local density of a point to its neighbors.
    \item \textbf{SO\_GAAL}: Single-Objective Generative Adversarial Active Learning \cite{sogaal}. SO\_GAAL generates adversarial samples to uncover anomalies in unsupervised settings.
    \item \textbf{LUNAR}: Unifying Local Outlier Detection Methods via Graph Neural Networks \cite{lunar}. LUNAR unifies and improves local outlier detection via graph neural networks.
    \item \textbf{VAE}: Variational AutoEncoder \cite{vae}. VAE uses reconstruction probabilities to detect anomalies.
    \end{itemize}
\end{itemize}


\begin{table*}[htb]
\setlength{\abovecaptionskip}{5pt}
\caption{Detailed information of five datasets used in \ours, including the original task, normal category(ies), anomaly category, the size of the training set, the size, and the anomaly ratio of the test set.}

\label{tab:appx_pre_dataset}
\centering
\setlength{\tabcolsep}{3.8pt}
\renewcommand{\arraystretch}{1.2}
\fontsize{8}{12}\selectfont 
\begin{tabular}{l l l l r r r}
\toprule
\textbf{Dataset} & \textbf{Original Task} &\textbf{ Normal Category(ies)} & \textbf{Anomaly Category} & \textbf{\# Train} & \textbf{\# Test} & \textbf{\% Anomaly} \\
\midrule
\textbf{AG News} & AG news topics classification & Sports, Business, Sci/Tech & World & 66,098 & 32,109 & 11.77\% \\
\textbf{BBC News} & BBC news topics classification & Business, Politics, Sport, Tech & Entertainment & 1,206 & 579 & 10.71\% \\
\textbf{IMDB Reviews} & \vspace{-3pt}binary sentiment classification & Positive & Negative & 17,417 & 8,952 & 16.61\% \\
 & of IMDb movie reviews & & & & & \\
\textbf{N24 News} & \vspace{-3pt}New York Times news & Television, Your Money, & Food & 40,569 & 19,227 & 9.51\% \\
 & classification & \vspace{-3pt}Automobiles, Science, & & & & \\
 & & \vspace{-3pt}Economy, Dance, Travel, & & & & \\
 & & \vspace{-3pt}Technology, Sports, Movies, & & & & \\
 & & \vspace{-3pt}Music, Real Estate, Books, & & & & \\
 & & \vspace{-3pt}Education, Art \& Design, & & & & \\
 & & \vspace{-3pt}Theater, Media, Style, & & & & \\
 & & \vspace{-3pt}Global Business, Well, & & & & \\
 & & \vspace{-3pt}Health, Fashion \& Style, & & & & \\
 & & Opinion  & & & & \\

\textbf{SMS Spam} & \vspace{-3pt}mobile phone SMS spam & Non-spam (Ham) & Spam & 3,162 & 1,510 & 10.20\% \\
 & messages detection & & & & & \\
\bottomrule
\end{tabular}
\vspace{-10pt}
\end{table*}

\section{Additional Details for Task 1}
\vspace{-0.5ex}
\label{appx:ad}
\subsection{Prompt Details}
\vspace{-0.3ex}
\label{appx:ad_prompt}
Prompt design is crucial for zero-shot LLM-based detection, as the performance heavily relies on its instructiveness and clarity.
As discussed in \S\ref{sec:ad_design} about LLM-based zero-shot AD, we evaluate two settings based on varying levels of prior knowledge in the real world: ``Normal Only'' and ``Normal + Anomaly.'' 
The LLM prompt template for setting ``Normal Only'' is provided in Table~\ref{tab:prompt_ad_normal_only}, and the prompt template for setting ``Normal + Anomaly'' is presented in Table~\ref{tab:prompt_ad_normal_anomaly}. 
The prompt templates of the two settings are different in the {\textbf{definition of anomaly}}, marked in \textcolor{Mahogany}{red} in Table~\ref{tab:prompt_ad_normal_anomaly}.

We utilize a series of prompt engineering techniques, including: 
\begin{itemize}[nosep, itemsep=3pt, leftmargin=*]
    \item \textit{Task Information} \cite{task_info}. It is essential to provide clear task information. We carefully define the detection scenario, the anomaly definition, and the rules to reduce hallucinations.
    \item \textit{Chain-of-Thought (CoT)} \cite{cot}. CoT prompting encourages LLMs to decompose their reasoning into sequential intermediate steps and organize information logically. We explicitly provide a completed chain of thoughts in the prompt.
    \item \textit{Explanation and Implicit CoT}. We require an explanation $r$ generated before the anomaly score $s$ for each inquiry as shown in Eq.~(\ref{eq:ad_prompt}). When generating the explanation, LLMs \textbf{implicitly} create the CoT in the background \cite{implicitly}. This approach aligns with the auto-regressive nature of decoder-only LLMs, encouraging them to think carefully and logically before determining the anomaly score, thereby enhancing reliability.
\end{itemize}

In our experiments, we discovered that Llama 3.1 requires implicit CoT. Presenting the anomaly score $s$ before the explanation $r$ causes the Llama 3.1-based detector to crash and consistently outputs $s=0$. This issue does not impact GPT-4o \add{and DeepSeek-V3}. We attribute this to their significantly larger parameter count, which grants it a stronger resilience to prompt changes.

\subsection{Complete Baseline Results}
\vspace{-0.2ex}
In addition to the top two baseline results in \S\ref{sec:ad_results}, we provide the complete results for all 18 baseline methods in Table~\ref{tab:appx_ad_baseline_results}. We observe that Llama 3.1 outperforms most of these baselines, further supporting the efficacy of zero-shot AD via LLMs shown in Table~\ref{tab:ad_results}. The baseline results presented in this study are derived from \citet{nlp_adbench}.



\subsection{Error Analysis}
\vspace{-0.2ex}
In our experiments, LLMs occasionally encountered errors. 
Llama 3.1 sometimes entered infinite loops, repeatedly generating the same sentence until reaching the token limit, as illustrated in the AG News example shown in Text Box~\ref{box:error_llama}, where the repetitive nonsensical response is highlighted in \textit{italics}. However, the underlying causes of such behavior remain underexplored.
\add{DeepSeek-V3 sometimes struggles to follow instructions, resulting in incorrect JSON formatting or failing to return any output. Additionally, it occasionally experiences internal server errors.}
For GPT-4o, outputs were sometimes restricted by OpenAI’s safety filters when addressing potentially harmful content, such as politically sensitive topics.

\begin{tcolorbox}[sharp corners, fontupper=\footnotesize, boxsep=1mm, left=1mm, right=1mm, top=1mm, bottom=1mm, title=An infinite loop example in Llama 3.1]
\label{box:error_llama}
\textbf{Text sample:} ``This week's TravelWatch column profiles Anangu Tours, an Aborigine owned tour company in Australia's Red Center.''\\
\textbf{Response:} \{"reason": "The text sample is about a tour company in Australia, which relates to travel and geography, making it most closely align with the Sports category is not the best fit, but it does not fit into the Business or Sci/Tech categories, but it does fit into the Sports category is not the best fit, but it does fit into the category of travel which is related to Sports, but more closely related to the category of travel which is not listed, \textit{but the closest is Sports, but it is more closely related to the category of travel which is not listed, but the closest is Sports, but it is more closely related to the category of travel which is not listed, but the closest is Sports, but it is more closely related to the category of travel which is not listed, but the closest is Sports...}
\end{tcolorbox}

\begin{table}[ht]
\setlength{\abovecaptionskip}{4pt}
\centering
\setlength{\tabcolsep}{4pt}
\renewcommand{\arraystretch}{0.9}
\fontsize{8}{12}\selectfont 
\begin{tabular}{l c c}
\toprule
\textbf{Dataset} & \textbf{``Normal Only''} &\textbf{``Normal + Anomaly''} \\
\midrule
\textbf{AG News} & 552 & 48 \\
\textbf{BBC News} & 0 & 3 \\
\textbf{IMDB Reviews} & 21 & 29 \\
\textbf{N24 News} & 299 & 898 \\
\textbf{SMS Spam} & 0 & 2 \\
\bottomrule
\end{tabular}
\caption{Error count in Llama 3.1}
\label{tab:appx_ad_error_llama}
\vspace{-3pt}
\end{table}

\begin{table}[ht]
\setlength{\abovecaptionskip}{4pt}
\centering
\setlength{\tabcolsep}{4pt}
\renewcommand{\arraystretch}{0.9}
\fontsize{8}{12}\selectfont 
\begin{tabular}{l c c}
\toprule
\textbf{Dataset} & \textbf{``Normal Only''} &\textbf{``Normal + Anomaly''} \\
\midrule
\textbf{AG News} & 1 & 0 \\
\textbf{BBC News} & 0 & 0 \\
\textbf{IMDB Reviews} & 1 & 9 \\
\textbf{N24 News} & 0 & 0 \\
\textbf{SMS Spam} & 0 & 0 \\
\bottomrule
\end{tabular}
\caption{Error count in GPT-4o}
\label{tab:appx_ad_error_gpt}
\vspace{-13pt}
\end{table}

\begin{table}[ht]
\setlength{\abovecaptionskip}{4pt}
\centering
\setlength{\tabcolsep}{4pt}
\renewcommand{\arraystretch}{0.9}
\fontsize{8}{12}\selectfont 
\begin{tabular}{l c c}
\toprule
\textbf{Dataset} & \textbf{``Normal Only''} &\textbf{``Normal + Anomaly''} \\
\midrule
\textbf{AG News} & 14 & 15 \\
\textbf{BBC News} & 3 & 1 \\
\textbf{IMDB Reviews} & 206 & 16 \\
\textbf{N24 News} & 138 & 252 \\
\textbf{SMS Spam} & 5 & 0 \\
\bottomrule
\end{tabular}
\caption{Error count in DeepSeek-V3}
\label{tab:appx_ad_error_deepseek}
\vspace{-13pt}
\end{table}


We exclude these errors from our results but provide the error counts in both ``Normal Only'' and ``Normal + Anomaly'' settings for reference in Table~\ref{tab:appx_ad_error_llama} for Llama 3.1,   Table~\ref{tab:appx_ad_error_gpt} for GPT-4o, \add{and Table~\ref{tab:appx_ad_error_deepseek} for DeepSeek-V3}.
Notably, the error counts vary between the two settings, suggesting that the triggers for errors, such as Llama’s infinite loop or GPT-4o’s safety filters, are sensitive to prompt variations. This occurs even though the prompts in both settings have similar semantic meanings.

\subsection{Verbal Score}
In the zero-shot AD task, we utilize LLM-generated verbal anomaly scores as a signal for detection. Verbalization methods are widely used because they offer an intuitive and straightforward estimation~\citep{xia2025survey}. However, LLMs can often be overconfident in their responses due to the influence of reinforcement learning from human feedback (RLHF)~\citep{kadavath2022language}.
\section{Additional Details for Task 2}
\label{appx:aug}
\subsection{Generating Synthetic Samples Details}
\label{appx:aug_syn}
\subsubsection{Evaluation Protocol and Prompt Details}
As discussed in \S\ref{subsec:aug_syn}, we set a scenario with limited training data $\mathcal{D}_{\text{small\_train}} = \left\{x_1, x_2, \dots, x_m\right\}$. Specifically, $\mathcal{D}_{\text{small\_train}}$ contains $v$ samples for each normal category $\mathcal{C}_{\text{normal}}^j \in \mathcal{C}_{\text{normal}} = \left\{\mathcal{C}_{\text{normal}}^1, \dots, \mathcal{C}_{\text{normal}}^k\right\}$, where $k$ is the number of normal categories.

We employ a multi-step strategy \add{with multiple rounds} to mitigate repetitive outputs, token limit constraints, and difficulties in handling long contexts. The detailed pipeline is outlined below:
\begin{enumerate}[nosep, itemsep=3pt, leftmargin=*]
    \item \textit{Keywords Generation.} To ensure a consistent synthetic data distribution compared with the original training data, $t$ groups of keywords are generated for each normal category $\mathcal{C}_{\text{normal}}^j$. We construct the prompt $\mathcal{P}_{\text{keywords}}$ using a template $T_{\text{keywords}}(\cdot)$ as shown in Table ~\ref{tab:prompt_aug_synth_keyword}. This template utilizes \texttt{\{name\}} and \texttt{\{original\_task\}} information from Table ~\ref{appx:pre_dataset}. The prompt $\mathcal{P}_{\text{keywords}}$ is processed by the LLM $f_{\text{LLM}}(\cdot)$ to produce $t \times k$ groups of keywords $\mathcal{K} = \left\{\mathcal{K}_1, \mathcal{K}_2, \dots, \mathcal{K}_{t \times k}\right\}$. Each keyword group $\mathcal{K}_i$ contains three keywords with increasing levels of granularity from coarse to fine.
    \item \textit{Synthetic Sample Generation.} We iterate the groups of keywords, constructing $\mathcal{P}_{\text{synth}}=\left\{\mathcal{P}_{\text{synth}}^1, \mathcal{P}_{\text{synth}}^2,\dots, \mathcal{P}_{\text{synth}}^{t\times k}\right\}$ using a template $T_{\text{synth}}(\cdot)$ as displayed in Table ~\ref{tab:prompt_aug_synth_sample}. Each prompt $\mathcal{P}_{\text{synth}}^j$ is fed into the LLM $f_{\text{LLM}}(\cdot)$ to generate a corresponding synthetic sample $\hat{x}_j$. Finally, we obtain a synthetic dataset $\mathcal{D}_{\text{synth}} = \left\{\tilde{x}_1, \tilde{x}_2, \dots, \tilde{x}_{t \times k}\right\}$.
\end{enumerate}
The pipeline is formally summarized as follows:
\begin{equation}
    \begin{aligned}
        \mathcal{P}_{\text{keywords}} &= T_{\text{keywords}}\bigl(\texttt{\{name\}},\texttt{\{original\_task\}}\bigr) \\
        \mathcal{K} &= f_{\text{LLM}}\left(\mathcal{P}_{\text{keywords}}\right) =\left\{\mathcal{K}_1, \dots, \mathcal{K}_{t\times k}\right\} \\
        \mathcal{P}_{\text{synth}}&=\left\{T_{\text{synth}}\left(\mathcal{K}_1\right),\dots, T_{\text{synth}}\left(\mathcal{K}_{t\times k}\right)\right\} \\
        \mathcal{D}_{\text{synth}}&= \left\{f_{\text{LLM}}\left(\mathcal{P}_{\text{synth}}^1\right), \dots, f_{\text{LLM}}\left(\mathcal{P}_{\text{synth}}^{t\times k}\right)\right\}\nonumber
        \label{eq:synth_process}
    \end{aligned}
\end{equation}

The prompt templates $T_{\text{keywords}}$ and $T_{\text{synth}}$ leverage the prompt techniques, including task information and CoT, as discussed in \S\ref{appx:ad_prompt}.

\subsubsection{Experiments Details and Challenges}
We set the number of samples from each normal category $\mathcal{C}_{\text{normal}}^j$ in the limited training set $\mathcal{D}_{\text{small\_train}}$ to $v=10$. Similarly, the number of synthetic samples generated for each normal category $\mathcal{C}_{\text{normal}}^j$ in the synthetic set $\mathcal{D}_{\text{synth}}$ is $t=50$ for the ``AG New'', ``BBC News'', ``IMDB Reviews'', and ``SMS Spam'' datasets. For the ``N24 News'' dataset, we set $v=3$ and $t=30$ due to its numerous normal categories.

We use GPT-4o for synthetic data generation. We observed that increasing $t$ occasionally causes them to terminate the keyword generation process before reaching the token limit. A similar issue occurs with Llama 3.1, even for smaller values of $t$. As a result, Llama 3.1 is excluded from this task. 
We presume these issues stem from the inherent challenges LLMs face in processing long contexts. 
\add{We also exclude DeepSeek-V3 due to its unsatisfactory results.}

\def\doubleunderline#1{\underline{\underline{#1}}}
\begin{table*}[htb]
\setlength{\abovecaptionskip}{4pt}
\caption{Performance comparison of AD baselines \textit{with} and \textit{without} LLM-generated synthetic data across five datasets. We also show the average performance $\pm$ its standard deviation over five datasets. The better results for each detector are highlighted in bold. The performance may vary due to the embedding changes.}
\label{tab:aug_synth_results}
\centering
\setlength{\tabcolsep}{2.1pt}
\renewcommand{\arraystretch}{0.95}
\fontsize{6.5}{9.5}\selectfont 
\begin{tabular}{r cc cc cc cc cc cc}
\toprule
\multicolumn{1}{c}{\multirow{2}{*}{\textbf{Training Set}}} & \multicolumn{2}{c}{\raisebox{0.3ex}{\textbf{AG News}}} & \multicolumn{2}{c}{\raisebox{0.3ex}{\textbf{BBC News}}} & \multicolumn{2}{c}{\raisebox{0.3ex}{\textbf{IMDB Reviews}}} & \multicolumn{2}{c}{\raisebox{0.3ex}{\textbf{N24 News}}} & \multicolumn{2}{c}{\raisebox{0.3ex}{\textbf{SMS Spam}}} & \multicolumn{2}{c}{\raisebox{0.3ex}{\textbf{Average Performance}}} \\ \cline{2-13}
\multicolumn{1}{c}{} & \raisebox{-0.3ex}{AUROC} \raisebox{-0.1ex}{$\uparrow$} & \raisebox{-0.3ex}{AUPRC} \raisebox{-0.1ex}{$\uparrow$} & \raisebox{-0.3ex}{AUROC} \raisebox{-0.1ex}{$\uparrow$} & \raisebox{-0.3ex}{AUPRC} \raisebox{-0.1ex}{$\uparrow$} & \raisebox{-0.3ex}{AUROC} \raisebox{-0.1ex}{$\uparrow$} & \raisebox{-0.3ex}{AUPRC} \raisebox{-0.1ex}{$\uparrow$} & \raisebox{-0.3ex}{AUROC} \raisebox{-0.1ex}{$\uparrow$} & \raisebox{-0.3ex}{AUPRC} \raisebox{-0.1ex}{$\uparrow$} & \raisebox{-0.3ex}{AUROC} \raisebox{-0.1ex}{$\uparrow$} & \raisebox{-0.3ex}{AUPRC} \raisebox{-0.1ex}{$\uparrow$} & \raisebox{-0.3ex}{AUROC} \raisebox{-0.1ex}{$\uparrow$} & \raisebox{-0.3ex}{AUPRC} \raisebox{-0.1ex}{$\uparrow$} \\ \midrule

\multicolumn{1}{c}{} & \multicolumn{12}{c}{\raisebox{0.3ex}{\textbf{OpenAI + AE}}} \\
\multicolumn{1}{r}{without $\mathcal{D}_{\text{synth}}$} & 0.5054 & 0.1189 & 0.6016 & 0.1309 & 0.5014 & 0.1665 & 0.7119 & 0.1681 & \textbf{0.5000} & \textbf{0.1020} & 0.5641$\pm$0.0834 & 0.1373$\pm$0.0262 \\
\multicolumn{1}{r}{with $\mathcal{D}_{\text{synth}}$} & \textbf{0.8097} & \textbf{0.3290} & \textbf{0.8434} & \textbf{0.3936} & \textbf{0.8097} & \textbf{0.3290} & \textbf{0.8097} & \textbf{0.3290} & 0.4341 & 0.0852 & \textbf{0.7413$\pm$0.1542} & \textbf{0.2932$\pm$0.1069} \\
\midrule
\multicolumn{1}{c}{} & \multicolumn{12}{c}{\raisebox{0.3ex}{\textbf{OpenAI + DeepSVDD}}} \\
\multicolumn{1}{r}{without $\mathcal{D}_{\text{synth}}$} & 0.5171 & 0.1237 & \textbf{0.6127} & \textbf{0.1415 }& \textbf{0.5667} & \textbf{0.1969} & \textbf{0.6278} & \textbf{0.1511} & \textbf{0.6398} & \textbf{0.1479} & \textbf{0.5928$\pm$0.0453} & \textbf{0.1522$\pm$0.0243} \\
\multicolumn{1}{r}{with $\mathcal{D}_{\text{synth}}$} & \textbf{0.5554} & \textbf{0.1365} & 0.5867 & 0.1267 & 0.5554 & 0.1365 & 0.5554 & 0.1365 & 0.3086 & 0.0681 & 0.5123$\pm$0.1026 & 0.1209$\pm$0.0267 \\
\midrule
\multicolumn{1}{c}{} & \multicolumn{12}{c}{\raisebox{0.3ex}{\textbf{OpenAI + ECOD}}} \\
\multicolumn{1}{r}{without $\mathcal{D}_{\text{synth}}$} & 0.5014 & 0.1180 & 0.5623 & 0.1208 & 0.5000 & 0.1661 & 0.6202 & 0.1311 & \textbf{0.4078} & \textbf{0.0789} & 0.5183$\pm$0.0709 & 0.1230$\pm$0.0279 \\
\multicolumn{1}{r}{with $\mathcal{D}_{\text{synth}}$} & \textbf{0.6709} & \textbf{0.1954} & \textbf{0.7660} & \textbf{0.3210} & \textbf{0.6709} & \textbf{0.1954} & \textbf{0.6709} & \textbf{0.1954} & 0.3351 & 0.0708 & \textbf{0.6228$\pm$0.1485} & \textbf{0.1956$\pm$0.0791} \\
\midrule
\multicolumn{1}{c}{} & \multicolumn{12}{c}{\raisebox{0.3ex}{\textbf{OpenAI + IForest}}} \\
\multicolumn{1}{r}{without $\mathcal{D}_{\text{synth}}$} & 0.6120 & 0.1620 & \textbf{0.7102} & 0.1903 & 0.5788 & 0.1947 & 0.5331 & 0.1010 & \textbf{0.6386} & \textbf{0.1467} & \textbf{0.6145$\pm$0.0594} & 0.1589$\pm$0.0340 \\
\multicolumn{1}{r}{with $\mathcal{D}_{\text{synth}}$} & \textbf{0.6759} & \textbf{0.2159} & 0.6655 & \textbf{0.2107} & \textbf{0.6759} & \textbf{0.2159} & \textbf{0.6759} & \textbf{0.2159} & 0.2700 & 0.0649 & 0.5926$\pm$0.1614 & \textbf{0.1847$\pm$0.0599} \\
\midrule
\multicolumn{1}{c}{} & \multicolumn{12}{c}{\raisebox{0.3ex}{\textbf{OpenAI + LOF}}} \\
\multicolumn{1}{r}{without $\mathcal{D}_{\text{synth}}$} & \textbf{0.6404} & \textbf{0.1661} & \textbf{0.7128} & \textbf{0.2565} & \textbf{0.6759} & \textbf{0.2485} & \textbf{0.7179} & \textbf{0.2061} & 0.7582 & 0.2445 & \textbf{0.7010$\pm$0.0400} & \textbf{0.2243$\pm$0.0339} \\
\multicolumn{1}{r}{with $\mathcal{D}_{\text{synth}}$} & 0.5469 & 0.1411 & 0.6513 & 0.2075 & 0.5469 & 0.1411 & 0.5469 & 0.1411 & \textbf{0.8150} & \textbf{0.2602} & 0.6214$\pm$0.1049 & 0.1782$\pm$0.0484 \\
\midrule
\multicolumn{1}{c}{} & \multicolumn{12}{c}{\raisebox{0.3ex}{\textbf{OpenAI + SO\_GAAL}}} \\
\multicolumn{1}{r}{without $\mathcal{D}_{\text{synth}}$} & \textbf{0.5657} & \textbf{0.1324} & \textbf{0.3240} & \textbf{0.0770} & \textbf{0.5388} & \textbf{0.1659} & 0.3351 & 0.0654 & \textbf{0.3953} & \textbf{0.0823} & \textbf{0.4318$\pm$0.1017} & \textbf{0.1046$\pm$0.0383} \\
\multicolumn{1}{r}{with $\mathcal{D}_{\text{synth}}$} &0.4461 & 0.0976 & 0.2787 & 0.0703 & 0.4461 & 0.0976 & \textbf{0.4461} & \textbf{0.0976} & 0.0698 & 0.0637 & 0.3374$\pm$0.1487 & 0.0854$\pm$0.0151 \\
\midrule
\multicolumn{1}{c}{} & \multicolumn{12}{c}{\raisebox{0.3ex}{\textbf{OpenAI + LUNAR}}} \\
\multicolumn{1}{r}{without $\mathcal{D}_{\text{synth}}$} & 0.6527 & 0.2035 & 0.8554 & 0.4670 & 0.6546 & 0.2315 & 0.7879 & 0.2473 & \textbf{0.1506} & \textbf{0.0573} & 0.6202$\pm$0.2475 & 0.2413$\pm$0.1314 \\
\multicolumn{1}{r}{with $\mathcal{D}_{\text{synth}}$} & \textbf{0.8651} & \textbf{0.4228} & \textbf{0.9330} & \textbf{0.7332} & \textbf{0.8651} & \textbf{0.4228} & \textbf{0.8651} & \textbf{0.4228} & 0.1375 & 0.0568 & \textbf{0.7332$\pm$0.2990} & \textbf{0.4117$\pm$0.2143} \\
\midrule

\multicolumn{1}{c}{} & \multicolumn{12}{c}{\raisebox{0.3ex}{\textbf{OpenAI + VAE}}} \\
\multicolumn{1}{r}{without $\mathcal{D}_{\text{synth}}$} & 0.6857 & 0.1842 & 0.7143 & 0.1816 & 0.5031 & 0.1670 & 0.6932 & 0.1698 & \textbf{0.5000} & \textbf{0.1020} & 0.6193$\pm$0.0966 & 0.1609$\pm$0.0302 \\
\multicolumn{1}{r}{with $\mathcal{D}_{\text{synth}}$} & \textbf{0.7905} & \textbf{0.3654} & \textbf{0.7674} & \textbf{0.2654} & \textbf{0.7905} & \textbf{0.3654} & \textbf{0.7905} & \textbf{0.3654} & 0.0696 & 0.0545 & \textbf{0.6417$\pm$0.2862} & \textbf{0.2832$\pm$0.1207} \\
\bottomrule
\end{tabular}
\vspace{-10pt}
\end{table*}

\add{We repeat the generation of keywords four times, with different temperatures $[1.0, 0.9, 0.8, 0.7]$ and different seeds $[42, 43, 44, 45]$. To further avoid repetition, we add additional sentences to the end of the prompts, including:
\begin{itemize}[nosep, itemsep=1pt, leftmargin=*]
\item \textit{``This is the first time you do this task, good luck!"}
\item \textit{``You've completed this task before, and you're improving at it."}
\item \textit{``After doing this task twice, you have a better understanding of it."}
\item \textit{``You have done this task three times, you are now an expert at it."}
\end{itemize}

\def\doubleunderline#1{\underline{\underline{#1}}}
\begin{table*}[htb]
\setlength{\abovecaptionskip}{5pt}
\caption{Complete model selection results across five datasets. We display the average AUROC and AUPRC of models recommended by querying each reasoning LLM five times (duplicates allowed). "Best Performance" marks the highest performance achieved by any baseline model for each dataset, while "Average Performance" denotes the mean performance across all baseline models.}
\label{tab:ums_synth_results}
\centering
\setlength{\tabcolsep}{2.5pt}
\renewcommand{\arraystretch}{0.95}
\fontsize{8}{12}\selectfont 
\begin{tabular}{r cc cc cc cc cc}
\toprule
\multicolumn{1}{c}{\multirow{2}{*}{\textbf{Settings}}} & \multicolumn{2}{c}{\raisebox{0.3ex}{\textbf{AG News}}} & \multicolumn{2}{c}{\raisebox{0.3ex}{\textbf{BBC News}}} & \multicolumn{2}{c}{\raisebox{0.3ex}{\textbf{IMDB Reviews}}} & \multicolumn{2}{c}{\raisebox{0.3ex}{\textbf{N24 News}}} & \multicolumn{2}{c}{\raisebox{0.3ex}{\textbf{SMS Spam}}} \\ \cline{2-11}
\multicolumn{1}{c}{} & \raisebox{-0.3ex}{AUROC} \raisebox{-0.1ex}{$\uparrow$} & \raisebox{-0.3ex}{AUPRC} \raisebox{-0.1ex}{$\uparrow$} & \raisebox{-0.3ex}{AUROC} \raisebox{-0.1ex}{$\uparrow$} & \raisebox{-0.3ex}{AUPRC} \raisebox{-0.1ex}{$\uparrow$} & \raisebox{-0.3ex}{AUROC} \raisebox{-0.1ex}{$\uparrow$} & \raisebox{-0.3ex}{AUPRC} \raisebox{-0.1ex}{$\uparrow$} & \raisebox{-0.3ex}{AUROC} \raisebox{-0.1ex}{$\uparrow$} & \raisebox{-0.3ex}{AUPRC} \raisebox{-0.1ex}{$\uparrow$} & \raisebox{-0.3ex}{AUROC} \raisebox{-0.1ex}{$\uparrow$} & \raisebox{-0.3ex}{AUPRC} \raisebox{-0.1ex}{$\uparrow$} \\ \midrule

\multicolumn{1}{l}{OpenAI-o1} & 0.7132 & 0.3199 & 0.6798 & 0.2831 & 0.6563 & 0.3278 & 0.7091 & 0.2419 & 0.3647 & 0.0752 \\
\multicolumn{1}{l}{OpenAI-o1-preview} & 0.8908 & 0.6193 & 0.6992 & 0.2214 & 0.6652 & 0.2787 & 0.7706 & 0.3422 & 0.5774 & 0.1220 \\
\multicolumn{1}{l}{OpenAI-o3-mini} & 0.6455 & 0.2401 & 0.7329 & 0.3132 & 0.5358 & 0.2521 & 0.4870 & 0.0952 & 0.5758 & 0.1169 \\
\multicolumn{1}{l}{DeepSeek-R1} & 0.8273 & 0.4744 & 0.7224 & 0.2424 & 0.7009 & 0.3976 & 0.7733 & 0.3113 & 0.5090 & 0.1022 \\
\midrule
\multicolumn{1}{l}{Baseline Average} & 0.6924 & 0.2685 & 0.7178 & 0.3574 & 0.5298 & 0.2038 & 0.6004 & 0.1585 & 0.5565 & 0.1277 \\
\multicolumn{1}{l}{Best Performance} & 0.9226 & 0.6918 & 0.9732 & 0.8653 & 0.7366 & 0.5165 & 0.8320 & 0.4425 & 0.7862 & 0.2450 \\
\bottomrule
\end{tabular}
\vspace{-6pt}
\end{table*}

\noindent We carefully examine and remove duplicate keyword groups. Out of 200 generations, there are typically fewer than 5 repeated groups, with a maximum of 15. It shows that our method is effective.
}

\subsubsection{Complete Results}
The detailed results are provided in Table \ref{tab:aug_synth_results}. These additional experiments on baselines follow the settings used in \citet{nlp_adbench}, except that the ``batch\_size'' is set $=4$ due to the amount of $\mathcal{D}_{\text{small\_train}}$ in AE, VAE, and DeepSVVD.


\subsubsection{Edges over LLM-based Zero-shot AD}
At first glance, LLM-based zero-shot AD could eliminate the need to generate synthetic datasets for traditional models. However, they address different needs and offer complementary advantages. 
LLM-based zero-shot detection requires no task-specific training, offering easy deployment, adaptability across scenarios, and real-time inference—ideal for dynamic environments. However, its high computational cost can limit scalability for long-term or large-scale use.

In contrast, LLM-generated synthetic data enables the training of traditional models, significantly reducing inference costs for long-term or high-frequency detection tasks. Moreover, synthetic data can be a valuable resource for fine-tuning LLMs \cite{xu2024wizardlm, mitra2023orca}. This dual utility highlights the importance of synthetic data generation as both a complementary and cost-efficient solution in the AD ecosystem.


\subsection{Genarating Category Description Details}
\label{appx:aug_desc}
\subsubsection{Prompt Details}
As discussed in \S\ref{subsec:aug_desc}, we generate category descriptions to enhance LLM-based zero-shot AD. The prompt template used for generating category descriptions is shown in Table ~\ref{tab:prompt_aug_desc}. It leverages the prompt techniques, including task information and CoT, as discussed in \S\ref{appx:ad_prompt}.

\subsubsection{A Universal Component}
LLM-generated category descriptions serve as a universal component that can be integrated into prompts to enhance any LLM-based task requiring category-specific information. 
In our study, we demonstrate its effectiveness in improving LLM-based zero-shot AD as shown in Table \ref{tab:aug_desc_results}. 
Additionally, these descriptions can enhance LLM-based synthetic data generation similarly. 
This approach aligns with the Native Chain-of-Thought (NCoT) process \cite{wang2024openr} used in OpenAI o1 \cite{gpto1}.
Extending this idea, other datasets with distinct structures could inspire the development of task-specific universal components, enabling tailored augmentation strategies for diverse LLM-based applications.


\section{Additional Details for Case Study 3}
\label{appx:ums}
\subsection{Evaluation Protocol and Prompt Details}
As discussed in \S\ref{subsec:aug_desc}, we utilize the information of both dataset and candidate models to achieve UMS. The prompt template used for generating category descriptions is shown in Table ~\ref{tab:prompt_aug_desc}. 

Importantly, we restrict our selection to two-step methods mentioned in \S\ref{appx:pre_baseline}, as the structural differences between end-to-end and two-step methods introduce additional complexities to an already challenging task.

\subsection{Failures on Popular LLMs}
Despite the promising results achieved with GPT-o1-preview, widely used LLMs like GPT-4o and Llama 3.1 struggle with zero-shot UMS, frequently recommending the same model regardless of dataset context. This limitation highlights the need for enhanced reasoning abilities to better analyze dataset-specific requirements, model strengths and weaknesses, and their overall compatibility.

\subsection{Complete Results}
The detailed results with precise numerical values are provided in Table~\ref{tab:ums_synth_results} for reference.

\begin{table*}[htb]
\setlength{\abovecaptionskip}{3pt}
\caption{Performance comparison of LLM-based detectors and baseline methods across five datasets. LLM-based detectors are evaluated under two settings as described in \S\ref{sec:ad_design} with AUROC and AUPRC as the metrics (higher (\raisebox{0.2ex}{$\uparrow$}), the better). 
The \textbf{best} results are highlighted in bold, the \doubleunderline{second-best} results are double-underlined, and the \underline{third-best} results are single-underlined.}

\label{tab:appx_ad_baseline_results}
\centering
\setlength{\tabcolsep}{2.1pt}
\renewcommand{\arraystretch}{0.95}
\fontsize{8}{12}\selectfont 
\begin{tabular}{c cc cc cc cc cc}
\toprule
\multicolumn{1}{c}{\multirow{2}{*}{\textbf{Settings}}} & \multicolumn{2}{c}{\raisebox{0.3ex}{\textbf{AG News}}} & \multicolumn{2}{c}{\raisebox{0.3ex}{\textbf{BBC News}}} & \multicolumn{2}{c}{\raisebox{0.3ex}{\textbf{IMDB Reviews}}} & \multicolumn{2}{c}{\raisebox{0.3ex}{\textbf{N24 News}}} & \multicolumn{2}{c}{\raisebox{0.3ex}{\textbf{SMS Spam}}} \\ \cline{2-11}
\multicolumn{1}{c}{} & \raisebox{-0.3ex}{AUROC} \raisebox{-0.1ex}{$\uparrow$} & \raisebox{-0.3ex}{AUPRC} \raisebox{-0.1ex}{$\uparrow$} & \raisebox{-0.3ex}{AUROC} \raisebox{-0.1ex}{$\uparrow$} & \raisebox{-0.3ex}{AUPRC} \raisebox{-0.1ex}{$\uparrow$} & \raisebox{-0.3ex}{AUROC} \raisebox{-0.1ex}{$\uparrow$} & \raisebox{-0.3ex}{AUPRC} \raisebox{-0.1ex}{$\uparrow$} & \raisebox{-0.3ex}{AUROC} \raisebox{-0.1ex}{$\uparrow$} & \raisebox{-0.3ex}{AUPRC} \raisebox{-0.1ex}{$\uparrow$} & \raisebox{-0.3ex}{AUROC} \raisebox{-0.1ex}{$\uparrow$} & \raisebox{-0.3ex}{AUPRC} \raisebox{-0.1ex}{$\uparrow$} \\ \midrule
\multicolumn{1}{c}{} & \multicolumn{10}{c}{\raisebox{0.3ex}{\textbf{Llama 3.1 8B Instruct}}} \\
\multicolumn{1}{l}{(1) with $\mathcal{C}_{\text{normal}}$} & 0.8226 & 0.4036 & 0.7910 & 0.3602 & 0.7373 & 0.3474 & 0.6267 & 0.1130 & 0.7558 & 0.2884\\
\multicolumn{1}{l}{(2) with $\mathcal{C}_{\text{normal}}$, $\mathcal{C}_{\text{anomaly}}$} & 0.8754 & 0.3998 & 0.8612 & 0.3960 & \underline{0.8625} & 0.4606 & \doubleunderline{0.8784} & \underline{0.3802} & \doubleunderline{0.9487} & \doubleunderline{0.6361} \\
\midrule
\multicolumn{1}{c}{} & \multicolumn{10}{c}{\raisebox{0.3ex}{\textbf{GPT-4o}}} \\
\multicolumn{1}{l}{(1) with $\mathcal{C}_{\text{normal}}$} & \textbf{0.9332} & \textbf{0.7207} & \underline{0.9574} & \underline{0.8432} & \doubleunderline{0.9349} & \doubleunderline{0.7823} & 0.7674 & 0.3252 & 0.7940 & 0.5568\\
\multicolumn{1}{l}{(2) with $\mathcal{C}_{\text{normal}}$, $\mathcal{C}_{\text{anomaly}}$} & \doubleunderline{0.9293} & \underline{0.6310} & \textbf{0.9919} & \textbf{0.9088} & \textbf{0.9668} & \textbf{0.8465} & \textbf{0.9902} & \textbf{0.9009} &\textbf{ 0.9862} & \textbf{0.8953}\\
\midrule
\multicolumn{1}{c}{{\textbf{\raisebox{0.3ex}{Methods}}}} & \multicolumn{10}{c}{\raisebox{0.3ex}{\textbf{Baselines}}} \\
CVDD & 0.6046 & 0.1296 & 0.7221 & 0.2976 & 0.4895 & 0.1576 & 0.7507 & 0.2886 & 0.4782 & 0.0712 \\
DATE & 0.8120 & 0.3996 & 0.9030 & 0.5764 & 0.5185 & 0.1682 & 0.7493 & 0.2794 & \underline{0.9398} & \underline{0.6112} \\
BERT + SO-GAAL & 0.4489 & 0.1033 & 0.3099 & 0.0849 & 0.4663 & 0.1486 & 0.4135 & 0.0837 & 0.3328 & 0.0714 \\
BERT + AE & 0.7200 & 0.2232 & 0.8839 & 0.4274 & 0.4650 & 0.1479 & 0.5749 & 0.1255 & 0.6918 & 0.1914 \\
BERT + DeepSVDD & 0.6671 & 0.2160 & 0.5683 & 0.1328 & 0.4287 & 0.1387 & 0.4366 & 0.0798 & 0.5859 & 0.1178 \\
BERT + ECOD & 0.6318 & 0.1616 & 0.6912 & 0.2037 & 0.4282 & 0.1374 & 0.4969 & 0.0928 & 0.5606 & 0.1156 \\
BERT + LOF & 0.7432 & 0.2549 & 0.9320 & 0.6029 & 0.4959 & 0.1621 & 0.6703 & 0.1678 & 0.7190 & 0.1837 \\
BERT + LUNAR & 0.7694 & 0.2717 & 0.9260 & 0.5943 & 0.4687 & 0.1497 & 0.6284 & 0.1436 & 0.6953 & 0.1817 \\
BERT + VAE & 0.6773 & 0.1878 & 0.7409 & 0.2559 & 0.4398 & 0.1405 & 0.4949 & 0.0957 & 0.6082 & 0.1360 \\
BERT + iForest & 0.6124 & 0.1559 & 0.6847 & 0.2131 & 0.4420 & 0.1412 & 0.4724 & 0.0872 & 0.5053 & 0.0994 \\
OpenAI + SO-GAAL & 0.5945 & 0.1538 & 0.2359 & 0.0665 & 0.6201 & 0.3005 & 0.5043 & 0.0963 & 0.5671 & 0.1213 \\
OpenAI + AE & 0.8326 & 0.4022 & 0.9520 & 0.7485 & 0.6088 & 0.1969 & 0.7155 & 0.1984 & 0.5511 & 0.1030 \\
OpenAI + DeepSVDD & 0.4680 & 0.1062 & 0.5766 & 0.1288 & 0.6563 & 0.3278 & 0.6150 & 0.1297 & 0.3491 & 0.0721 \\
OpenAI + ECOD & 0.7638 & 0.3294 & 0.7224 & 0.2424 & 0.7366 & \underline{0.5165} & 0.7342 & 0.2238 & 0.4317 & 0.0821 \\
OpenAI + LOF & 0.8905 & 0.5443 & 0.9558 & 0.7714 & 0.6156 & 0.2133 & 0.7806 & 0.2248 & 0.7862 & 0.2450 \\
OpenAI + LUNAR & \underline{0.9226} & \doubleunderline{0.6918} & \doubleunderline{0.9732} & \doubleunderline{0.8653} & 0.6474 & 0.2193 & \underline{0.8320} & \doubleunderline{0.4425} & 0.7189 & 0.1640 \\
OpenAI + VAE & 0.8144 & 0.3659 & 0.7250 & 0.2424 & 0.4515 & 0.1486 & 0.7418 & 0.2537 & 0.4259 & 0.0812 \\
OpenAI + iForest & 0.5213 & 0.1278 & 0.6064 & 0.1376 & 0.5064 & 0.1724 & 0.4944 & 0.0913 & 0.3751 & 0.0772 \\
\bottomrule
\end{tabular}
\vspace{-12pt}
\end{table*}

\begin{table*}[htb]
\setlength{\abovecaptionskip}{3pt}
\caption{LLM prompt template used for zero-shot AD in ``Normal Only'' setting discussed in \S\ref{sec:ad_design}. \texttt{\{normal\_category\_$x$\}} refers to the name of $x_{th}$ normal category. \texttt{\{text\}} represents the test sample to be detected.}
\label{tab:prompt_ad_normal_only}
\centering
\fontsize{10}{12}\selectfont 
\begin{tabular}{p{0.95\textwidth}}
\toprule
You are an intelligent and professional assistant that detects anomalies in text data.\\
\#\# Task:\\
- Following the rules below, determine whether the given text sample is an anomaly. Provide a brief explanation of your reasoning and assign an anomaly confidence score between 0 and 1.\\
\\
\#\# Categories:\\
- **\texttt{\{normal\_category\_1\}}**\\
- **\texttt{\{normal\_category\_2\}}**\\
- ...\\
\\
\#\# Rules:\\
1. **Anomaly Definition**:\\
\ \ \ \ - A text sample is considered an **anomaly** if it does **not** belong to **any of the categories** listed above.\\
2. **Scoring**:\\
\ \ \ \ - Assign an anomaly confidence score between 0 and 1.\\
\ \ \ \ - Use higher scores when you are highly confident in your decision.\\
\ \ \ \ - Use lower scores when you are uncertain or think the text sample is **not** an anomaly.\\
3. **Step-by-step Reasoning** (Chain of Thought):\\
\ \ \ \ - **Step 1**. Read the entire text sample carefully and understand it thoroughly.\\
\ \ \ \ - **Step 2**. Analyze the text sample by comparing its content to each category listed in the "Categories" section above, considering factors such as main topics, meanings, background, sentiments, etc.\\
\ \ \ \ - **Step 3**. Determine which category the text sample **most closely aligns with**.\\
\ \ \ \ \ \ \ \ - If it aligns with any category, it is **not** an anomaly.\\
\ \ \ \ \ \ \ \ - If it does **not** align with any category, it is an anomaly.\\
\ \ \ \ - **Step 4**. Assign an anomaly confidence score based on how confident you are that the text sample is an anomaly.\\
4. **Additional Notes**:\\
\ \ \ \ - A text sample may relate to multiple categories, but it should be classified into the **most relevant** one in this task.\\
\ \ \ \ - If you are uncertain whether the text sample **significantly aligns** with **any of the anomaly category(ies)**, assume that it does **not**, which means it is **not** an anomaly.\\
5. **Response Format**:\\
\ \ \ \ - Provide responses in a strict **JSON** format with the keys "reason" and "anomaly\_score."\\
\ \ \ \ \ \ \ \ - "reason": Your brief explanation of the reasoning in one to three sentences logically.\\
\ \ \ \ \ \ \ \ - "anomaly\_score": Your anomaly confidence score between 0 and 1.\\
\ \ \ \ - Ensure the JSON output is correctly formatted, including correct placement of commas between key-value pairs.\\
\ \ \ \ - Add a backslash (\textbackslash) before any double quotation marks (") within the values of JSON output for proper parsing (i.e., from " to \textbackslash"), and ensure that single quotation marks (') are preserved without escaping.\\
Text sample:\\
"\texttt{\{text\}}"\\
\\
Response in JSON format:\\
\\
\bottomrule
\end{tabular}
\end{table*}
\begin{table*}[htb]
\setlength{\abovecaptionskip}{3pt}
\caption{LLM prompt template used for zero-shot AD in ``Normal + Anomaly'' setting discussed in \S\ref{sec:ad_design}. \texttt{\{normal\_category\_$x$\}} refers to the name of $x_{th}$ normal category and \texttt{\{anomaly\_category\}} refers to the name of anomaly category. \texttt{\{text\}} represents the test sample to be detected. The different part compared with the prompt in the ``Normal Only'' setting is marked in \textcolor{Mahogany}{red}.}
\label{tab:prompt_ad_normal_anomaly}
\centering
\fontsize{10}{12}\selectfont 
\begin{tabular}{p{0.95\textwidth}}
\toprule
You are an intelligent and professional assistant that detects anomalies in text data.\\
\#\# Task:\\
- Following the rules below, determine whether the given text sample is an anomaly. Provide a brief explanation of your reasoning and assign an anomaly confidence score between 0 and 1.\\
\\
\#\# Categories:\\
\textcolor{Mahogany}{\#\#\# Normal Category(ies):}\\
- **\texttt{\{normal\_category\_1\}}**\\
- **\texttt{\{normal\_category\_2\}}**\\
- ...\\
\textcolor{Mahogany}{\#\#\# Anomaly Category(ies):}\\
\textcolor{Mahogany}{- \texttt{\{anomaly\_category\}}}\\
\\
\#\# Rules:\\
1. **Anomaly Definition**:\\
\ \ \ \ - \textcolor{Mahogany}{A text sample is considered an **anomaly** if it belongs to the **anomaly category(ies)** rather than **any of the normal category(ies)** listed above.}\\
2. **Scoring**:\\
\ \ \ \ - Assign an anomaly confidence score between 0 and 1.\\
\ \ \ \ - Use higher scores when you are highly confident in your decision.\\
\ \ \ \ - Use lower scores when you are uncertain or think the text sample is **not** an anomaly.\\
3. **Step-by-step Reasoning** (Chain of Thought):\\
\ \ \ \ - **Step 1**. Read the entire text sample carefully and understand it thoroughly.\\
\ \ \ \ - **Step 2**. Analyze the text sample by comparing its content to each category listed in the "Categories" section above, considering factors such as main topics, meanings, background, sentiments, etc.\\
\ \ \ \ - **Step 3**. Determine which category the text sample **most closely aligns with**.\\
\ \ \ \ \ \ \ \ - \textcolor{Mahogany}{If it **most closely aligns with** **any of the anomaly category(ies)**, it is an **anomaly**.}\\
\ \ \ \ \ \ \ \ - \textcolor{Mahogany}{If it **most closely aligns with** **any of the normal category(ies)** instead, it is **not** an anomaly.}\\
\ \ \ \ - **Step 4**. Assign an anomaly confidence score based on how confident you are that the text sample is an anomaly.\\
4. **Additional Notes**:\\
\ \ \ \ - A text sample may relate to multiple categories, but it should be classified into the **most relevant** one in this task.\\
\ \ \ \ - If you are uncertain whether the text sample **significantly aligns** with **any of the anomaly category(ies)**, assume that it does **not**, which means it is **not** an anomaly.\\
5. **Response Format**:\\
\ \ \ \ - Provide responses in a strict **JSON** format with the keys "reason" and "anomaly\_score."\\
\ \ \ \ \ \ \ \ - "reason": Your brief explanation of the reasoning in one to three sentences logically.\\
\ \ \ \ \ \ \ \ - "anomaly\_score": Your anomaly confidence score between 0 and 1.\\
\ \ \ \ - Ensure the JSON output is correctly formatted, including correct placement of commas between key-value pairs.\\
\ \ \ \ - Add a backslash (\textbackslash) before any double quotation marks (") within the values of JSON output for proper parsing (i.e., from " to \textbackslash"), and ensure that single quotation marks (') are preserved without escaping.\\
Text sample:\\
"\texttt{\{text\}}"\\
\\
Response in JSON format:\\
\\
\bottomrule
\end{tabular}
\end{table*}
\begin{table*}[htb]
\setlength{\abovecaptionskip}{3pt}
\caption{LLM prompt template used for keyword generation, which is the first step of generating synthetic samples as discussed in \S\ref{subsec:aug_syn}. \texttt{\{normal\_category\_$x$\}} refers to the name of $x_{th}$ normal category. \texttt{\{name\}} and \texttt{\{original\_task\}} can be found in Tab.~\ref{appx:pre_dataset}. \texttt{\{num\_keyword\_groups\}} set the number of keyword groups that LLM needs to generate for each category.}
\label{tab:prompt_aug_synth_keyword}
\centering
\fontsize{10}{12}\selectfont 
\begin{tabular}{p{0.95\textwidth}}
\toprule
You are an intelligent and professional assistant that generates groups of keywords for given categories in a dataset.\\
\#\# Task:\\
- Following the rules below, generate **exactly**  \texttt{\{num\_keyword\_groups\}} unique keyword groups for **each given category** according to your understanding of the category (and its description).\\
- Each keyword group will be used to generate synthetic data for the corresponding category.\\
\\
\#\# Rules:\\
1. **Keyword Group Generation**:\\
\ \ \ \ - For **each given category**, generate **exactly** \texttt{\{num\_keyword\_groups\}} keyword groups. Each group should contain exactly three keywords, with different levels of granularity: one broad/general, one intermediate, and one fine-grained.\\
\ \ \ \ - Ensure that the three keywords in each group are thematically related to each other and align with the category's description.\\
\ \ \ \ - Avoid redundancy or overly similar keywords across different groups.\\
\ \ \ \ - Ensure that each group is unique and relevant to the key topics described in the category.\\
2. **Granularity**:\\
\ \ \ \ - The first keyword should be broad/general, representing a high-level or overarching topic.\\
\ \ \ \ - The second keyword should be intermediate, more specific than the first, but not overly narrow.\\
\ \ \ \ - The third keyword should be fine-grained and specific, related to detailed subtopics or precise aspects of the category.\\
3. **Response Format**:\\
\ \ \ \ - For each given category, provide the keyword groups as a list, where each entry is a group of three keywords (broad, intermediate, fine-grained).\\
\ \ \ \ - Structure the response so that the key is the category name, and the value is a list of generated keyword groups.\\
\ \ \ \ - Ensure the JSON output is properly formatted, including correct placement of commas between key-value pairs and no missing brackets.\\
\ \ \ \ - Add a backslash (\textbackslash) before any double quotation marks (") within the values of JSON output for proper parsing (i.e., from " to \textbackslash"), and ensure that single quotation marks (') are preserved without escaping.\\
\\
The "\texttt{\{name\}}" dataset's original task is \texttt{\{original\_task\}}. It contains the following category(ies):\\
\texttt{\{normal\_category\_1\}}\\
\texttt{\{normal\_category\_2\}}\\
...\\
\\
Response in JSON format:\\
\\
\bottomrule
\end{tabular}
\end{table*}
\begin{table*}[htb]
\setlength{\abovecaptionskip}{3pt}
\caption{LLM prompt template used for sample generation, which is the second step of generating synthetic samples as discussed in \S\ref{subsec:aug_syn}. We generate a single synthetic sample per keyword group. \texttt{\{keyword\_group[$i$]\}} refers to ${(i+1)}_{th}$ granularity level's keyword in this keyword group. \texttt{\{category\}} represents the name of the corresponding category for this keyword group.}
\label{tab:prompt_aug_synth_sample}
\centering
\fontsize{10}{12}\selectfont 
\begin{tabular}{p{0.95\textwidth}}
\toprule
You are an intelligent and professional assistant that generates a synthetic text sample based on a group of 3 keywords with different levels of granularity.\\
\#\# Task:\\
- Generate a synthetic text sample that incorporates the provided group of 3 keywords (broad, intermediate, and fine-grained) listed below.\\
- The generated sample should align with the meanings and themes suggested by the keywords provided.\\
\\
\#\# Rules:\\
1. **Sample Characteristics**:\\
\ \ \ \ - Generate a synthetic text sample that naturally incorporates the three provided keywords (broad, intermediate, and fine-grained).\\
\ \ \ \ - Ensure that the text sample is coherent and contextually relevant to the themes suggested by the keywords.\\
2. **Keyword Usage**:\\
\ \ \ \ - The three keywords must appear naturally within the content.\\
\ \ \ \ - Ensure that the broad keyword sets the overall context, the intermediate keyword refines the discussion, and the fine-grained keyword offers more detailed insight into a specific subtopic.\\
3. **Response Format**:\\
\ \ \ \ - Provide the generated sample as a single string response representing the text sample.\\
\ \ \ \ - Ensure the output is in a readable format.\\
\ \ \ \ - Do not include any additional messages or commentary.\\
\ \ \ \ - Add a backslash (\textbackslash) before any double quotation marks (") within the values of JSON output for proper parsing (i.e., from " to \textbackslash"), and ensure that single quotation marks (') are preserved without escaping.\\
\\
The "\texttt{\{name\}}" dataset's original task is \texttt{\{original\_task\}}. The category is "\texttt{\{category\}}", and the group of keywords to use is:\\
- Broad: \texttt{\{keyword\_group[0]\}}\\
- Intermediate: \texttt{\{keyword\_group[1]\}}\\
- Fine-grained: \texttt{\{keyword\_group[2]\}}\\
\\
Response in JSON format:\\
\\
\bottomrule
\end{tabular}
\end{table*}
\begin{table*}[htb]
\setlength{\abovecaptionskip}{3pt}
\caption{LLM prompt template for generating category descriptions discussed in \S\ref{subsec:aug_desc}. \texttt{\{normal\_category\_$x$\}} refers to the name of $x_{th}$ normal category and \texttt{\{anomaly\_category\}} refers to the name of anomaly category. \texttt{\{name\}} and \texttt{\{original\_task\}} can be found in Tab.~\ref{appx:pre_dataset}.}
\label{tab:prompt_aug_desc}
\centering
\fontsize{10}{12}\selectfont 
\begin{tabular}{p{0.95\textwidth}}
\toprule
You are an intelligent and professional assistant that generates descriptions for given categories in a text dataset.\\
\#\# Task:\\
- Following the rules below, generate detailed textual descriptions that explain the main characteristics, typical topics, and common examples for each given category.\\
\\
\#\# Rules:\\
1. For each category, provide a continuous, coherent description in a single paragraph that includes:\\
\ \ \ \ - **Definition or overview**: Start by briefly defining or describing the category in one to two sentences. If you list multiple aspects or features in the definition (such as related fields or industries), ensure you append expressions like "etc." or "and so on" to indicate that the list is not exhaustive.\\
\ \ \ \ - **Main topics or subjects**: Highlight the typical topics or subjects covered by this category. Ensure that you use phrases like "etc." or "and so on" at the end of each list to indicate that the list is not exhaustive.\\
	- **Relevant examples**: Mention examples of content that belong to this category. Also, use expressions like "etc." or "and so on" at the end of the list to show that these are illustrative, not exhaustive.\\
2. Use **step-by-step reasoning** to ensure the descriptions are logical and clear.\\
3. Each description should be clear, coherent, and helpful for someone unfamiliar with the dataset and the task.\\
4. Always append phrases like "etc." or "and so on" to lists or enumerations of examples, topics, or aspects, **including the definition part**.\\
5. Response Format:\\
\ \ \ \ - Provide a response where each key is the category name, and the value is the corresponding description as a continuous paragraph.\\
\ \ \ \ - Ensure the JSON output is correctly formatted, including correct placement of commas between key-value pairs.\\
\ \ \ \ - Add a backslash (\textbackslash) before any double quotation marks (") within the values of JSON output for proper parsing (i.e., from " to \textbackslash"), and ensure that single quotation marks (') are preserved without escaping.\\
\\
The "\texttt{\{name\}}" dataset's original task is \texttt{\{origianl\_task\}}. It contains the following categories:\\
\texttt{\{normal\_category\_1\}}\\
\texttt{\{normal\_category\_2\}}\\
...\\
\texttt{\{anomaly\_category\}}\\
\\
Response in JSON format:\\
\\
\bottomrule
\end{tabular}
\end{table*}
\begin{table*}[!ht]
\vspace{-0.2in}
\setlength{\abovecaptionskip}{3pt}
\caption{LLM prompt template used for UMS discussed in \S\ref{sec:ums}. \texttt{\{normal\_category\_$x$\}} refers to the name of $x_{th}$ normal category and \texttt{\{anomaly\_category\}} refers to anomaly one. We randomly select examples from the training set for both normal and anomaly data, denoted as \texttt{\{normal\_text\}} and \texttt{\{anomaly\_text\}}. \texttt{\{name\}}, \texttt{\{size\}} (i.e., \# of test set), and \texttt{\{original\_task\}} can be found in Tab.~\ref{appx:pre_dataset}. \texttt{\{avg\_len\}}, \texttt{\{max\_len\}}, \texttt{\{min\_len\}}, and \texttt{\{std\_len\}} are statistics of datasets as shown in Tab.~\ref{tab:appx_data_stat}. \texttt{\{abstract\}} is the abstract in the published paper of each model.}
\label{tab:prompt_ums}
\centering
\fontsize{10}{12}\selectfont 
\begin{tabular}{p{0.95\textwidth}}
\toprule
You are an expert in model selection for anomaly detection on text datasets.\\
\\
\#\# Task:\\
- Given the information of a dataset and a set of models, select the model you believe will achieve the best performance for detecting anomalies in this dataset. Provide a brief explanation of your choice.\\
\\
\#\# Dataset Information:\\
- Dataset Name: \texttt{\{name\}}\\
- Dataset Size: \texttt{\{size\}}\\
- Background: This dataset is originally for \texttt{\{original\_task\}}.\\
- Data Structure: Textual data with multiple categories. One category is considered anomalous, while the others are normal.\\
\ \ \ \ - Normal Category(ies): \texttt{\{normal\_category\_1\}}, \texttt{\{normal\_category\_2\}}\\
\ \ \ \ \ \ \ \ - An Example: \texttt{\{normal\_text\}}\\
\ \ \ \ - Anomaly Category: \texttt{\{anomaly\_category\}}\\
\ \ \ \ \ \ \ \ - An Example: \texttt{\{anomaly\_text\}}\\
-  Text Length Statistics:\\
\ \ \ \ - Average Length: \texttt{\{avg\_len\}}\\
\ \ \ \ - Maximum Length: \texttt{\{max\_len\}}\\
\ \ \ \ - Minimum Length: \texttt{\{min\_len\}}\\
\ \ \ \ - Standard Deviation: \texttt{\{std\_len\}}\\
\\
\#\# Model Information:\\
- Models utilize language models to generate embeddings and feed the embeddings into the models.\\
- We provide the abstracts of the papers that introduce the models for your reference.\\
\#\#\# Model Options:\\
- AutoEncoder (AE): \texttt{\{abstract\}} \cite{ae}\\
- Deep Support Vector Data Description (DeepSVDD): \texttt{\{abstract\}} \cite{deepsvdd}\\
- Empirical-Cumulative-Distribution-Based Outlier Detection (ECOD):\texttt{\{abstract\}} \cite{ecod}\\
- Isolation Forest (IForest): \texttt{\{abstract\}} \cite{iforest}\\
- Local Outlier Factor (LOF): \texttt{\{abstract\}} \cite{lof}\\
- Unifying Local Outlier Detection Methods via Graph Neural Networks (LUNAR): \texttt{\{abstract\}} \cite{lunar}\\
- Single-Objective Generative Adversarial Active Learning (SO-GAAL): \texttt{\{abstract\}} \cite{sogaal}\\
- Variational AutoEncoder (VAE): \texttt{\{abstract\}} \cite{vae}\\
\#\#\# Embedding Options:\\
- Bidirectional Encoder Representations from Transformers (BERT): \texttt{\{abstract\}} \cite{bert}\\
- "text-embedding-3-large" from OpenAI (referred to as OpenAI): \texttt{\{abstract\}} \cite{openai2024embedding}\\
\\
\#\# Rules:\\
1. Availabel options include "BERT+AE", "BERT+DeepSVDD", "BERT+ECOD", "BERT+iForest", "BERT+LOF", "BERT+LUNAR", "BERT+SO-GAAL", "BERT+VAE", "OpenAI+AE", "OpenAI+DeepSVDD", "OpenAI+ECOD", "OpenAI+iForest", "OpenAI+LOF", "OpenAI+LUNAR", "OpenAI+SO-GAAL", "OpenAI+VAE."\\
2. Treat all models equally and evaluate them based on their compatibility with the dataset characteristics and the anomaly detection task.\\
3. Response Format:\\
\ \ \ \ - Provide responses in a strict **JSON** format with the keys "reason" and "choice."\\
\ \ \ \ \ \ \ \ - "reason": Your explanation of the reasoning.\\
\ \ \ \ \ \ \ \ - "choice": The model you have selected for anomaly detection in this dataset.\\
\\
Response in JSON format:\\
\\
\bottomrule
\end{tabular}
\end{table*}

\end{document}